%% file: main.tex
\documentclass[10pt,twocolumn,letterpaper]{article}

\usepackage[pagenumbers]{cvpr} 

\input{preamble}
\definecolor{cvprblue}{rgb}{0.21,0.49,0.74}
\usepackage[pagebackref,breaklinks,colorlinks,allcolors=cvprblue]{hyperref}
\usepackage{multirow}
\usepackage{makecell}
\usepackage{xcolor}
\usepackage{algorithm}
\usepackage{algorithmic}
\usepackage{cuted}
\usepackage{float}
\makeatletter
\renewcommand\paragraph{%
  \@startsection{paragraph}{4}{\z@}%
  {1pt}
  {-1em}
  {\normalfont\normalsize\bfseries}}
\makeatother

\newtheorem{theorem}{Theorem}
\newtheorem{definition}{Definition}

\def\paperID{17979} 
\def\confName{CVPR}
\def\confYear{2026}

\title{Universal Guideline-Driven Image Clustering via a Hybrid LLM Agent}

\author{Wenliang Zhong$^1$, Rob Barton$^2$, Lucas Goncalves$^2$, Kushal Kumar$^2$, Feng Jiang$^1$, \\ Hehuan Ma$^1$, Yuzhi Guo$^1$, Vidit Bansal$^2$, Karim Bouyarmane$^2$, and Junzhou Huang$^1$ \\
$^1$The University of Texas at Arlington, $^2$Amazon\\
{\tt\small \{wxz9204, fxj8843, hehuan.ma, yuzhi.guo\}@mavs.uta.edu}\\
{\tt\small \{rab, sglucas, kushlku, bansalv, bouykari\}@amazon.com} \\
{\tt\small jzhuang@uta.edu}
}

\begin{document}
\maketitle
\begin{strip}
    \centering
    \vspace{-1.8cm}  \includegraphics[width=1.0\linewidth]{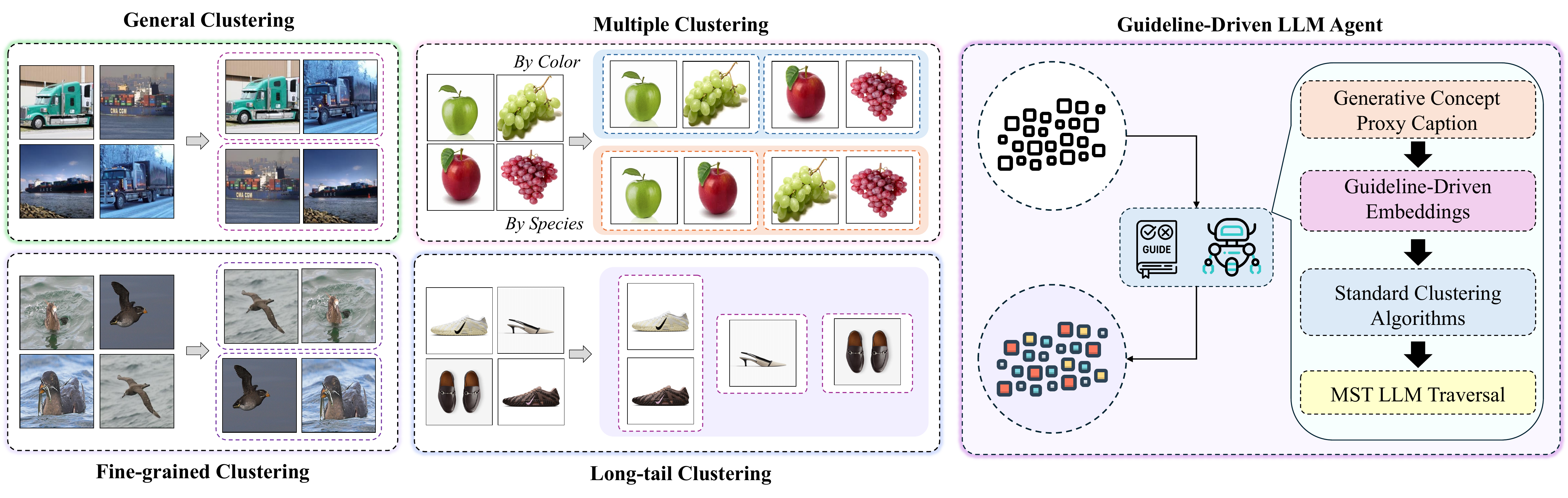}
    \vspace{-0.8cm}
    \captionof{figure}
    {\textbf{Overview of our Guideline-Driven Clustering Agent}. We introduce the first universal clustering framework that handles diverse image clustering scenarios through textual guidelines, spanning from general to fine-grained tasks, from global to local criteria, and from balanced to long-tail distributions. Our training-free hybrid agent flexibly adapts across these diverse clustering requirements.}
    \label{fig: paper_overview}
\end{strip}

\input{sec/0_abstract}    
\input{sec/1_intro}

\input{sec/2_formatting}
\input{sec/3_finalcopy}
{
    \small
    \bibliographystyle{ieeenat_fullname}
    \bibliography{main}
}
\input{sec/X_suppl}

\end{document}

%% file: sec/0_abstract.tex
\begin{abstract}
Unifying image clustering across different clustering scenarios remains challenging due to fundamental gaps among tasks. We introduce a Guideline-Driven Image Clustering Agent, the first universal framework that bridges these gaps through textual guidelines. To incorporate complex guidelines without task-specific training, we propose Generative Concept Proxy Modeling, which generates guideline-aware embeddings via concept proxy extraction. 
For scenarios requiring automatic cluster discovery, we introduce LLM Traversal based on Minimum Spanning Tree that selectively applies LLM reasoning for complex semantic judgments. Our method generalizes across diverse clustering scenarios spanning from general to fine-grained categorization, from global to local criteria, and from balanced to long-tail distributions. 
Our framework consistently outperforms specialized methods across diverse clustering tasks. Project page is at: https://clustering-agent.github.io/.
\end{abstract}

%% file: sec/1_intro.tex
\vspace{-0.5cm}
\section{Introduction}
\label{sec:intro}
Image clustering, the task of grouping visually similar images into coherent categories, is fundamental to numerous applications ranging from content organization to visual search systems. Traditional clustering approaches rely on static encoders to generate embeddings and use clustering algorithms such as K-Means~\cite{macqueen1967multivariate, hartigan1979algorithm} and DBSCAN~\cite{ester1996density}, which fundamentally rely on mathematical distance metrics in feature spaces rather than semantic understanding of visual content. Recent deep clustering methods~\cite{ji2019invariant, qian2023stable, liu2024interactive, lilearning} have introduce specific training strategies to steer the clustering process. However, a critical gap persists: methods optimized for general object categorization fail at fine-grained distinctions, approaches designed for balanced distributions struggle with long-tail data, and techniques for single-perspective clustering cannot handle multiple simultaneous criteria. Most fragmentation necessitates task-specific solutions, limiting practical deployment of clustering systems across diverse requirements.

To address this fundamental limitation, we aim to introduce the first universal clustering framework driven by textual guidelines. Our approach accepts natural language descriptions specifying clustering criteria, from simple directives ("group by color") to complex multi-attribute requirements ("organize athletic footwear by brand and intended activity"). This guideline-driven paradigm unifies diverse clustering tasks spanning general to fine-grained categorization, global to local criteria, and balanced to long-tail distributions under a single framework.

Existing text-guided approaches~\cite{braverman2025learningaugmented, hao2025task,nguyen2025forensic, kalavas2025towards, chakrabarti2025globallocal, wangunified, qi2024control, pham2025rapid, li2023image} face fundamental limitations in processing such guidelines. They handle only single, concrete criteria at a time~\cite{yao2024multi, yao2024customized} (e.g., group by color OR species, not both), require task-specific training for new criteria~\cite{geng2025personalized, chen2025agent}, or assume predetermined cluster numbers~\cite{nguyen2025forensic}. These constraints prevent handling composite criteria and abstract semantic reasoning essential for real-world applications. 

To incorporate complex, multi-faceted guidelines without task-specific training, one intuitive solution is to use instruction-aware embedders~\cite{su2022one, wang2023improving, zhang2024gme} and treat guidelines as instructions. However, direct encoding of images with guidelines faces significant challenges. First, existing multimodal instruction-aware embedders~\cite{zhang2024gme, na2025segment, zhang2023multimodal} fail to process images following complex guidelines. Second, they may cause important attributes in the guideline to be overshadowed by visually dominant but irrelevant features. 
As a solution, we propose Generative Concept Proxy Modeling (GCPM).
GCPM addresses this through a two-stage process, by first extracting concept proxy captions via multimodal large language models (MLLMs), then encoding these concept proxies using instruction-aware embedders. This achieves attribute disentanglement and enables guideline composition within embeddings that traditional models fail to capture. The resulting embeddings support standard clustering algorithms, e.g., K-Means with cluster number prior or HDBSCAN~\cite{campello2013density} for automatic discovery.

For automatic clusters discovery, existing algorithms provide limited similarity comparison based on embeddings. While recent Large Language Models~\cite{TheC3, bai2025qwen2} (LLMs) can provide sophisticated reasoning capabilities for understanding complex guidelines and making sophisticated similarity judgments, existing approaches~\cite{zhang2023clusterllmlargelanguagemodels,viswanathan2023largelanguagemodelsenable,liu2025llm,fu2025mark,lin2025spill, jo2024zerodl} show critical computational concerns using LLMs. 
To address this challenge while preserving semantic reasoning capabilities, we introduce an LLM Traversal algorithm based on Minimum Spanning Tree (MST). 
This hybrid design leverages embedding-based efficiency for routine clustering decisions while applying selective LLM reasoning only where semantic complexity demands it.

The main contributions of this work are: 
(1) We introduce the first universal clustering framework capable of handling composite and abstract textual guidelines without requiring task-specific training, enabling flexible clustering according to complex, multi-criteria requirements. 
(2) We propose GCPM, a training-free embedding approach that efficiently incorporate guideline into embeddings that previous methods fail to achieve. GCPM captions augment critical attributes of images for better representation and disentanglement. 
(3) We present MST-based LLM traversal, a novel algorithm that significantly minimizes expensive LLM invocations by intelligently selecting which clustering decisions require deep semantic reasoning. 
(4) We provide comprehensive evaluation across four distinct clustering tasks, demonstrating the superior generalization and performance of our approach.

%% file: sec/2_formatting.tex
\section{Related Works}
\label{sec: related}

\paragraph{General Deep Clustering} Methods like~\cite{lilearning, liu2024interactive} propose specific training strategies for unsupervised learning. Though many of them  achieve impressive performance in general clustering, they require dedicated training and are often difficult to extend to different scenarios.

\paragraph{Text-guided and Multiple Clustering}
Methods have emerged to address semantic clustering needs. Recent advances include Multi-Sub~\cite{yao2024multi} and Multi-MaP~\cite{yao2024customized}, which enable user-specified perspectives through proxy learning approaches. TAC~\cite{li2023image} leverages external concepts to enhance general clustering.
However, they are fundamentally limited to single-criteria clustering or constrained to specific scenarios due to their embedding strategies.

\paragraph{Clustering Methods based on LLMs}
Existing methods leverage language models' semantic capabilities but face efficiency challenges. ClusterLLM~\cite{zhang2023clusterllmlargelanguagemodels} guides clustering through triplet comparisons yet requires embedding fine-tuning and handles only single criteria. While these approaches~\cite{zhang2023clusterllmlargelanguagemodels, viswanathan2023largelanguagemodelsenable,liu2025llm,fu2025mark,lin2025spill, jo2024zerodl} demonstrate LLMs' potential, they suffer from computational inefficiency requiring extensive model invocations that limit practical deployment. IC$|$TC~\cite{kwon2023image} pioneered training-free image clustering pipelines via LLMs but remains constrained to single, concrete criteria and costly iterations over the dataset.
 
\paragraph{Fine-grained Deep Clustering}
Methods~\cite{viswanathan2023largelanguagemodelsenable,kim2021contrastive,benny2020onegan} like DiFiC~\cite{du2024informationtheoreticgenerativeclusteringdocuments} achieve superior performance through diffusion model semantic extraction in fine-grained clustering. However, these require dataset-specific training and lack mechanisms for incorporating user-specified guidelines.

Our work differs fundamentally by processing composite criteria under various scenarios, achieving semantic sophistication and training-free efficiency through a hybrid approach. We carefully discuss our difference in Appendix~\ref{sec: compare_difference}.

%% file: sec/3_finalcopy.tex
\section{Methodology}
\begin{figure*}[!th]
    \centering
    \includegraphics[width=1.0\linewidth]{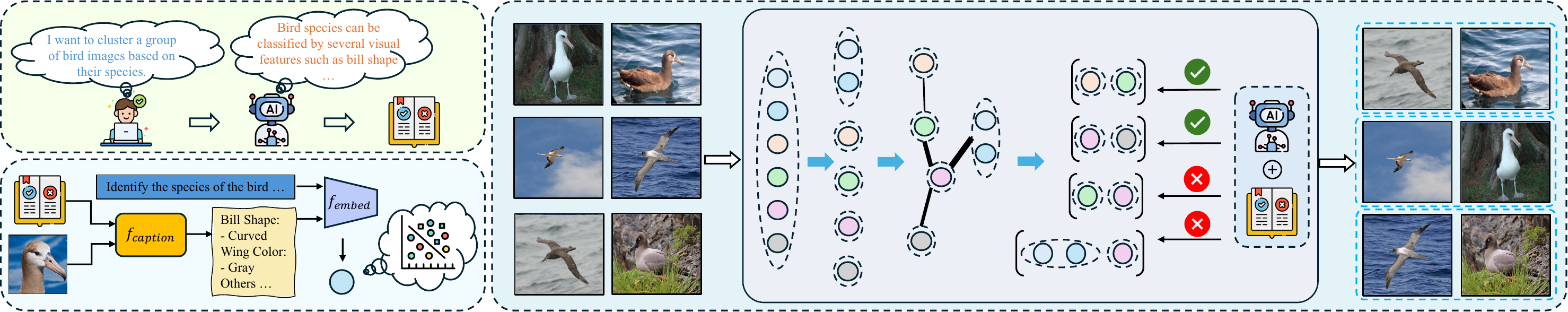}
    \caption{\textbf{Overview of the Clustering Framework}.  \textbf{Bottom left}: Generative Concept Proxy Modeling extracts guideline-aware textual descriptions from images via multimodal LLM and encodes them into embeddings for efficient clustering; \textbf{Right}: MST-based LLM Traversal refines initial clusters by constructing a Minimum Spanning Tree and selectively querying the LLM for semantic merging decisions. \textbf{Top left}: For scenarios without available guidelines, we introduce heuristic prompts for automatic guideline generation.}
    \label{fig: method_overview}
\vspace{-0.4cm}
\end{figure*}
Our Guideline-driven clustering agentic framework addresses the challenge of clustering images according to complex textual guidelines.
As shown in Figure \ref{fig: method_overview}, the framework operates through a two-stage hybrid approach that balances semantic understanding with computational efficiency, enabling training-free generalization across diverse clustering tasks. The first stage, Generative Concept Proxy Modeling, generates guideline-aware embeddings suitable for efficient clustering using traditional algorithms. The second stage employs MST-based LLM-guided Traversal to handle complex semantic judgments that require deep understanding of abstract concepts.

\subsection{Preliminaries}
\label{sec: preliminaries}
\paragraph{Problem Formulation}
Given a guideline $G$ and a set of samples $X = \{x_1, x_2, \cdots, x_N\}$, our objective is to find a function $f$ that groups samples into clusters $\mathcal{C} = \{C_1, C_2, \cdots, C_M\}$, where each cluster $C_i = \{x^i_1, x^i_2, \cdots\}$ contains semantically similar samples according to the specified guideline:
\begin{equation}
    \mathcal{C} = f(G, X).
\end{equation}
The guideline $G$ can be either directly provided by users such as standard operating procedures in business problems or derived from LLM-based knowledge through our heuristic prompting inspired by data substructure exploration.

\paragraph{Guideline Processing}
We assume guidelines are in textual form, containing multiple criteria that users wish to consider for clustering. Formally, a guideline $G$ contains a set of grouping attributes $A = \{a_1, a_2, \cdots, a_k\} \subseteq G$, where $A$ represents the universal set of possible attributes. For instance, in bird species clustering~\cite{WahCUB_200_2011}, $G$ might include attributes such as ``tail shape" and ``wing color". In scenarios where explicit guidelines are unavailable, we leverage the internal knowledge of LLMs~\cite{TheC3} to generate appropriate clustering guidelines. We introduce a series of heuristic prompts (shown in Appendix~\ref{sec: guideline_gen}) that require only basic understanding of the clustering objectives from users. For datasets where users have limited prior knowledge, unsupervised data substructure exploration techniques~\cite{luo2024llm, liu2024organizing} can be employed to identify fundamental criteria, which are then fed to the LLM for guideline generation. In our experiments, we evaluate performance on both datasets with predefined guidelines and those requiring LLM-generated guidelines.

\subsection{Generative Concept Proxy Modeling}
The Generative Concept Proxy Modeling (GCPM) generates guideline-aware embeddings suitable for efficient clustering. GCPM aims to address fundamental challenges in guideline-driven clustering: (1) combining textual guidelines with visual input to produce semantically meaningful representations requires careful consideration of how attributes in the guideline interact; (2) clustering tasks demand flexible dimensional focus: comprehensive tasks like species clustering~\cite{WahCUB_200_2011, KhoslaYaoJayadevaprakashFeiFei_FGVC2011, Nilsback08} require holistic consideration of all attributes, while multiple clustering tasks~\cite{yao2024multi, yao2024customized} like poker card categorization~\cite{gpiosenka_cardsdataset} on suits need emphasis on specific dimensions; (3) achieving generalization across domains without dataset-specific training remains challenging for existing deep clustering approaches~\cite{geng2025personalized, lilearning}.

To address these challenges, we leverage recent instruction-aware embedding models~\cite{wang2023improving, jiang2024e5, zhong2025zero, zhong2024compositional, zhou2025homie} that can incorporate textual instructions during the encoding process. These models offer the potential for training-free guideline adherence by directly encoding images alongside clustering instructions. However, directly encoding images and guidelines using multimodal instruction-aware embedding models~\cite{jiang2024vlm2vec, lin2024mm, liu2025lamra, zhang2024gme} introduces critical limitations that prevent effective clustering. In comprehensive clustering scenarios, fine-grained attributes mentioned in guidelines may be overshadowed if they are not explicitly queried. Moreover, in dimension-specific clustering, irrelevant but visually salient attributes can dominate the intended attributes. 

We address this through a concept proxy approach. We first employ a multimodal LLM (MLLM)~\cite{bai2025qwen2} as a captioning model $f_{caption}$ to extract concept-focused descriptions:
\begin{equation}
    c_i = f_{caption}(A, x_i), A \subseteq G.
\end{equation}
This process explicitly surfaces guideline-specified attributes in textual form, achieving attribute disentanglement. Subsequently, we use an instruction-aware embedding model $f_{embed}$ to encode the concept proxy caption:
\begin{equation}
    h_i = f_{embed}(S, c_i), S \subseteq G,
\end{equation}
where $S$ specifies the clustering focus required. It can be a single attribute or a global definition of all attributes in $A$. 
The resulting embeddings can be directly applied to standard clustering algorithms such as K-Means:
\begin{equation}
    \mathcal{C} = \text{Clustering}(H),
\end{equation}
where $H = \{h_1, h_2,\cdots, h_N\}$. This training-free approach enables cross-domain generalization while outperforming supervised methods on public benchmarks.


\begin{algorithm}[h]
\small
\caption{MST-based LLM Traversal}
\fontsize{8.8pt}{9.5pt}\selectfont
\begin{algorithmic}[1]
\REQUIRE Initial clusters $C^{(0)} = \{C_1, C_2, \ldots, C_M\}$ from HDBSCAN, Singleton points $S$, Guideline $G$, LLM merging function $f_{\text{merge}}$, Embeddings $H$
\ENSURE Refined clusters $C = \{C_1, C_2, \ldots, C_K\}$
\STATE $C \gets C^{(0)} \cup \{\{s\} \mid s \in S\}$
\STATE $M \gets |C|$ // Total number of clusters
\WHILE{$M > 1$}
    \STATE Compute Ward distance matrix $D \in \mathbb{R}^{M \times M}$ for all cluster pairs using Eq. (6)
    \STATE $T \gets \text{MST}(D)$ // Construct minimum spanning tree
    \STATE Sort edges in $T$ by distance in ascending order
    \STATE $\text{merged} \gets \text{False}$
    \FOR{each edge $(C_i, C_j) \in T$ in sorted order}
        \IF{$(C_i, C_j)$ has been rejected in previous iterations}
            \STATE \textbf{continue} // Skip pairs queried and rejected before
        \ENDIF
        \STATE $p \gets f_{\text{merge}}(G, C_i, C_j)$ // Query LLM for decision. Each cluster is represented by GCPM captions of samples.
        \IF{$p = 1$}
            \STATE $C_{\text{new}} \gets C_i \cup C_j$ // Merge clusters
            \STATE $C \gets (C \setminus \{C_i, C_j\}) \cup \{C_{\text{new}}\}$
            \STATE Update subsequent edges containing $C_i$ or $C_j$
            \STATE $M \gets M - 1$
            \STATE $\text{merged} \gets \text{True}$
        \ENDIF
    \ENDFOR
    \IF{$\text{merged} = \text{False}$}
        \STATE \textbf{break} // No clusters merged in this iteration
    \ENDIF
\ENDWHILE
\RETURN $C$
\end{algorithmic}
\label{algo: mst_traversal}
\end{algorithm}
\vspace{-1.0\baselineskip}

\subsection{MST-based LLM Traversal}
\label{sec: mst_traversal}
GCPM provides a training-free and efficient embedding approach for clustering. For scenarios where cluster numbers are predetermined, K-Means can be directly applied. However, in many real-world tasks, clusters are automatically discovered following guideline specifications rather than predefined knowledge. An intuitive solution is to apply algorithms that do not require predefined cluster numbers, such as HDBSCAN~\cite{campello2013density}, to  guideline-aware embeddings.
\begin{equation}
    \mathcal{C} = \text{HDBSCAN}(H).
\end{equation}
In practice, we observe that HDBSCAN tends to generate consistent small clusters but fails to merge homogeneous clusters into larger groups. It indicates that HDBSCAN effectively groups clearly similar samples together. But for scenarios involving larger sample volumes or more complex sample variations, it is less effective due to the density-based nature. To address this limitation, we propose using LLMs to merge these small clusters. LLMs can effectively understand complex guidelines and measure sample similarity at the semantic level, enabling more precise clustering decisions that go beyond embedding-based metrics.

However, LLM-based approaches~\cite{zhang2023clusterllmlargelanguagemodels,viswanathan2023largelanguagemodelsenable,liu2025llm,fu2025mark,lin2025spill, jo2024zerodl} are computationally expensive and time-consuming. Assuming HDBSCAN produces $M$ clusters, a naive approach would require $O(M^2)$ LLM comparisons to determine which clusters should be merged. To improve efficiency, we propose a Minimum Spanning Tree (MST) based traversal algorithm that strategically reduces the number of required LLM queries while maintaining clustering quality.

\paragraph{Ward Distance Computation}
For all clusters generated by HDBSCAN, we compute pairwise distances using Ward distance~\cite{ward1963hierarchical}, which measures the increase in sum of squares when merging two clusters $C_1$ and $C_2$:
\begin{equation}
\begin{aligned}
d(C_1, C_2) 
&= \sum_{i \in C_1 \cup C_2} \left\| h_i - m_{C_1 \cup C_2} \right\|^2 \\
&\quad - \sum_{i \in C_1} \left\| h_i - m_{C_1} \right\|^2 
- \sum_{i \in C_2} \left\| h_i - m_{C_2} \right\|^2 \\
&= \frac{|C_1| \cdot |C_2|}{|C_1| + |C_2|}
\left\| m_{C_1} - m_{C_2} \right\|^2,
\end{aligned}
\end{equation}
where $m_{C_i}$ is the centroid of the cluster $C_i$. This distance metric is particularly suitable for hierarchical clustering as it considers both cluster size and centroid separation.

\paragraph{MST Construction and Traversal:}
We construct an MST $T$ from the distance matrix $D\in \mathbb{R}^{M\times M}$:
\begin{equation}
    T = \text{MST}(D).
\end{equation}
The MST provides a traversal order that prioritizes cluster pairs with smallest distances, ensuring that the most promising merge candidates are evaluated first by the LLM.

\paragraph{Iterative Guideline-Driven LLM Merging}
Instead of merging clusters recursively until all are combined, we iteratively query an LLM to decide whether pairs of clusters should be merged. In each iteration, after computing the distance matrix and MST, we present cluster pairs to the merging LLM~\cite{bai2025qwen2} $f_{merge}$ in order of increasing distance:
\begin{equation}
    p = f_{merge}(G,C_i, C_j), \text{for each edge } (C_i, C_j) \in T.
\end{equation}
The LLM evaluates a cluster pair based on the provided guideline G and generates a binary decision $p \in \{0,1\}$ whether they should be merged. Each cluster is represented by GCPM captions of Top-K (K=5) samples nearest to the centroid. The process continues until either all appropriate clusters are merged or the LLM decides no further merging is beneficial. To further improve efficiency, we cache LLM decisions from previous iterations to avoid redundant queries for previously evaluated pairs. 
We discuss its efficiency improvement in the Appendix~\ref{sec: runtime_complexity}.

This hybrid design leverages embedding-based efficiency for initial clustering while applying selective LLM reasoning for complex semantic judgments. The complete process is in Algorithm \ref{algo: mst_traversal}. The MST Traversal optimally allocates LLM resources by prioritizing the most promising merge candidates. Moreover, it naturally supports \textit{incremental clustering}~\cite{charikar1997incremental} by treating new samples as individual clusters and merging them with existing structures, while many traditional algorithms require re-clustering.
\paragraph{Theoretical Complexity Analysis}
We provide theoretical justification for the efficiency of our MST-based approach. Based on empirical observations and reasonable assumptions about merge probability patterns, we prove that the expected number of LLM calls is $O(M \log M)$, compared to $O(M^2)$ for naive pairwise comparison methods. The complete proof is provided in Appendix~\ref{sec: proof_complexity}.

\subsection{ Long-Tail E-Commerce Clustering}
To evaluate our approach on challenging realistic guideline-driven scenarios, we introduce a dataset that reflects the long-tail clustering challenges commonly encountered in e-commerce. The data is derived from the Amazon Berkeley Objects (ABO) dataset~\cite{collins2022abo}, utilizing product images and textual attributes such as brand, manufacturer, and material.

\begin{figure}
    \centering
    \includegraphics[width=1.0\linewidth]{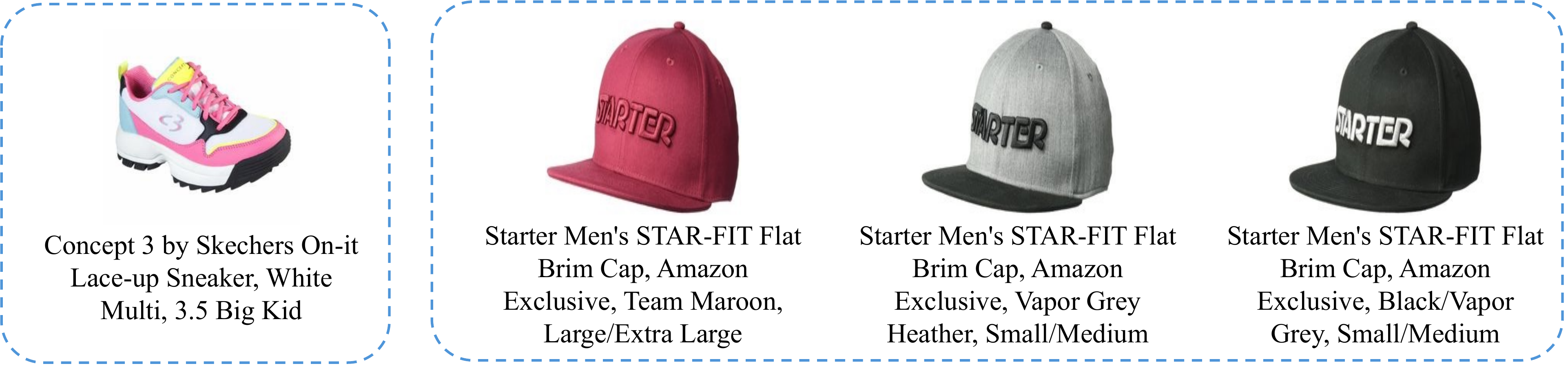}
    \vspace{-0.5cm}
    \caption{Examples from ABO-LC. Each item is represented by an image and an item name. Different clusters have various sizes.}
    \label{fig: abo_example}
\vspace{-0.5cm}
\end{figure}

We perform grouping based on meaningful attribute combinations that reflect real-world product organization practices. The resulting dataset exhibits the long-tail distribution characteristic of commercial platforms, where numerous specialized products form singleton or very small clusters. Examples are in Figure~\ref{fig: abo_example}. In summary, the dataset contains 10,756 products organized into 4,952 ground truth clusters, with 78.7\% of clusters containing two or fewer samples. We provide more statistics and discuss how data is processed, filtered, and grouped in Appendix~\ref{sec: abo_processing}.

The long-tail characteristic makes traditional clustering algorithms less effective, as methods like K-Means assume balanced cluster sizes and struggle with extreme imbalance scenarios. The dataset serves as a crucial benchmark for evaluating clustering methods on realistic scenarios where cluster numbers are unknown and sizes vary dramatically. 

\section{Experiments}
\subsection{Experimental Settings}
\begin{table}[t]
\centering
\caption{Statistics of clustering datasets and their criteria.}
\vspace{-0.2cm}
\setlength{\tabcolsep}{3.8pt}
\scalebox{0.7}{
\begin{tabular}{lccc}
\toprule
Dataset & Type & Criteria & \# of Samples \\
\midrule
CIFAR-10~\cite{krizhevsky2009learning} & GC & common objects & 60,000 \\
STL-10~\cite{coates2011analysis} & GC & common objects & 13,000 \\
ImageNet-10~\cite{deng2009imagenet} & GC & common objects & 13,000 \\
Fruit~\cite{hu2017finding} & MC & fruit color; fruit species & 105 \\
Cards~\cite{yao2023augdmc} & MC & card number; card suits & 8,029 \\
CIFAR10-MC~\cite{yao2024customized} & MC & object types; object environment & 60,000 \\
CUB Birds~\cite{WahCUB_200_2011} & FC & bird species & 5,794 \\
Stanford Cars~\cite{krause20133d} & FC & car types & 8,041 \\
Stanford Dogs~\cite{KhoslaYaoJayadevaprakashFeiFei_FGVC2011} & FC & dog breeds & 8,580 \\
Oxford Flowers~\cite{Nilsback08} & FC & flower species & 6,149 \\
ABO\cite{collins2022abo}-LC (Proposed) & LC & e-commerce items & 10,756 \\
\bottomrule
\end{tabular}
}
\label{tab: dataset_stats}
\vspace{-0.3cm}
\end{table}
\paragraph{Tasks and Datasets}
We conduct experiments on a wide range of image clustering tasks, including general clustering (GC), multiple clustering (MC), fine-grained clustering (FC), and our proposed long-tail e-commerce clustering (LC). We select commonly used datasets for experiments: STL-10~\cite{coates2011analysis}, ImageNet-10~\cite{deng2009imagenet}, and CIFAR-10~\cite{krizhevsky2009learning} for GC; CIFAR10-MC~\cite{yao2024customized}, Fruit~\cite{hu2017finding}, and Cards~\cite{yao2023augdmc} for MC; CUB Birds~\cite{WahCUB_200_2011}, Stanford Dogs~\cite{KhoslaYaoJayadevaprakashFeiFei_FGVC2011}, Stanford Cars~\cite{krause20133d}, Oxford Flowers~\cite{Nilsback08} for FC; and the proposed ABO~\cite{collins2022abo}-LC. Because evaluation metrics are different among tasks, we follow existing baselines to use metrics and process datasets in each task. The dataset statistic is shown in Table \ref{tab: dataset_stats}. We use Accuracy (ACC), Normalized Mutual Information (NMI), and Adjusted Rand Index (ARI) for GC; NMI and Rand Index (RI) for MC; ACC and NMI for FC; ACC, NMI, and ARI for LC. We discuss details of datasets, their guideline availability, and metrics in Appendix~\ref{sec: data_and_metrics}. All LLM prompts, including guideline, are in Appendix~\ref{sec: detailed_guideline}. 
Note that we maintain an unsupervised setting because none of our prompts includes ground-truth labels.

\paragraph{Models}
We use QWen2.5-VL-Instruct~\cite{bai2025qwen2} (7B) for both generating concept proxy caption and MST Traversal though the merging process in traversal only requires texts. For embeddings, we validate our method on three distinct embedders to test the robustness: INSTRUCTOR-large~\cite{su2022one} (335M), a regular instruction-aware text encoder; E5-Mistral~\cite{wang2023improving} (7B), an LLM-based encoder; GME-Qwen2-VL~\cite{zhang2024gme} (7B), an MLLM-based encoder. We use only captions for INSTRUCTOR and E5-Mistral while both captions and images for GME-Qwen. Detailed forwarding processes of embedders are discussed in Appendix~\ref{sec: forward_process}. All models are run in inference mode without further training. Experiments are validated separately on a server with NVIDIA A6000 and a server with H100 GPUs, providing the ease for adapting to new tasks, scenarios, and data.

\begin{table}[!tbp]
\centering
\caption{Performance comparison on general clustering. 
Baselines assume known number of clusters. 
KMS.: K-Means; HDBS.: HDBSCAN. Max $\Delta\uparrow$ measures improvements of MST Traversal.}
\vspace{-0.2cm}
\setlength{\tabcolsep}{3.5pt}
\scalebox{0.7}{
\begin{tabular}{lccc|ccc|ccc}
\toprule
\multirow{2}{*}{Method} & \multicolumn{3}{c|}{CIFAR-10} & \multicolumn{3}{c|}{STL-10} & \multicolumn{3}{c}{ImageNet-10} \\
\cmidrule(lr){2-4} \cmidrule(lr){5-7} \cmidrule(lr){8-10}
& ACC & NMI & ARI & ACC & NMI & ARI & ACC & NMI & ARI \\
\midrule
Cop-KMS.~\cite{wagstaff2001constrained}  & 89.0 & 82.3 & 78.6 & 85.4 & 78.1 & 73.1 & 88.6 & 85.5 & 81.0 \\
IIC~\cite{ji2019invariant}              & 61.7 & 51.3 & 41.1 & 49.9 & 43.1 & 29.5 & -    & -    & -    \\
DCCM~\cite{wu2019deep}                  & 62.3 & 49.6 & 40.8 & 48.2 & 37.6 & 26.2 & 71.0 & 60.8 & 55.5 \\
BYOL~\cite{grill2020bootstrap}          & 83.3 & 72.3 & 68.0 & 80.7 & 68.0 & 63.5 & 82.1 & 73.2 & 67.4 \\
MiCE~\cite{tsai2020mice}                & 83.5 & 73.7 & 69.8 & 75.2 & 63.5 & 57.5 & 90.1 & 84.2 & 82.2 \\
PCL~\cite{li2020prototypical}           & 87.4 & 80.2 & 76.6 & 81.0 & 71.8 & 67.0 & 90.7 & 84.1 & 82.2 \\
PICA~\cite{huang2020deep}               & 64.5 & 56.1 & 46.7 & -    & -    & -    & 85.0 & 78.2 & 73.3 \\
SCAN~\cite{van2020scan}                 & 88.3 & 79.7 & 77.2 & 80.9 & 69.8 & 64.6 & -    & -    & -    \\
FixMatch~\cite{sohn2020fixmatch}        & 92.8 & 86.8 & 85.4 & 68.6 & 61.7 & 49.2 & 92.5 & 84.2 & 84.4 \\
CC~\cite{li2021contrastive}             & 79.0 & 70.5 & 63.7 & 85.0 & 76.4 & 72.6 & 89.3 & 85.9 & 82.2 \\
GCC~\cite{zhong2021graph}               & 85.6 & 76.4 & 72.8 & 78.8 & 68.4 & 63.1 & 90.7 & 84.1 & 82.2 \\
IDFD~\cite{tao2021clustering}           & 81.5 & 71.4 & 66.6 & 74.1 & 62.3 & 55.0 & 86.2 & 75.5 & 69.0 \\
NMM~\cite{dang2021nearest}             & 84.3 & 74.8 & 70.9 & 80.8 & 69.4 & 65.0 & -    & -    & -    \\
TCC~\cite{shen2021you}                  & 90.6 & 79.0 & 73.3 & 81.4 & 73.2 & 68.9 & 89.7 & 84.8 & 82.0 \\
ProPos~\cite{huang2022learning}         & 92.0 & 85.6 & 84.1 & 83.2 & 72.1 & 70.4 & 89.0 & 81.2 & 78.7 \\
SPICE~\cite{niu2022spice}               & 83.8 & 73.4 & 70.5 & 90.8 & 81.7 & 81.2 & 92.1 & 82.8 & 83.6 \\
TCL~\cite{li2022twin}                   & 88.9 & 82.2 & 78.4 & 88.6 & 82.0 & 78.5 & 90.4 & 89.6 & 85.0 \\
CoNR~\cite{yu2023contextually}          & 85.3 & 78.1 & 71.1 & 81.2 & 70.2 & 69.2 & 87.7 & 79.2 & 75.0 \\
DMICC~\cite{li2023dual}                 & 83.1 & 72.2 & 67.5 & 75.5 & 64.1 & 60.5 & 92.5 & 84.3 & 85.2 \\
SeCu~\cite{qian2023stable}              & 93.0 & 86.1 & 85.7 & 83.6 & 73.3 & 69.3 & -    & -    & -    \\
DivClust~\cite{metaxas2023divclust}     & 81.9 & 72.4 & 68.1 & -    & -    & -    & 93.6 & 89.1 & 87.8 \\
IC$|$TC~\cite{kwon2023image} & 88.4 & 78.9 & 75.9 & 97.4 & 93.9 & 94.4 & - & - & - \\
IDCTCL~\cite{liu2024interactive}        & 92.7 & 84.4 & 84.8 & 92.7 & 85.3 & 84.6 & 97.2 & 93.2 & 93.9 \\
LFSS~\cite{lilearning}                  & 93.4 & 87.2 & \textbf{86.6} & 86.1 & 77.1 & 74.0 & 93.2 & 85.6 & 85.7 \\
\midrule
\multicolumn{4}{l|}{\textit{KMS. w/ Known \# of Clus.}}  &  &  &    &  &  \\
GCPM-I (Ours) & 75.6 & 66.6 & 60.1 & 98.2 & 95.5 & 96.0 & 97.5 & 92.8 & 86.4 \\
GCPM-E (Ours) & 82.6 & 69.8 & 65.9 & 97.9 & 95.0 & 95.4 & 97.1 & 94.4 & 93.9 \\
GCPM-G (Ours) & \textbf{94.1} & \textbf{87.5} & 84.5 & \textbf{98.8} & \textbf{96.9} & \textbf{97.4} & \textbf{98.8} & \textbf{96.7} & \textbf{97.1} \\
\midrule
\multicolumn{4}{l|}{\makecell[l]{\textit{HDBS. \& Our MST Traversal} \\ \textit{ w/o Known \# of Clus.}}}&  &  &  &  &  &  \\
GCPM-I (Ours) & - & 40.1 & 0.3 & - & 43.6 & 0.5 & - & 41.3 & 0.2 \\
+MST Traversal & - & 53.6 & 43.9 & - & 59.1 & 28.9 & - & 75.5 & 72.3 \\
GCPM-E (Ours) & - & 41.4 & 0.3 & - & 44.1 & 0.7 & - & 41.8 & 0.3 \\
+MST Traversal & - & 59.3 & 56.0 & - & 59.5 & 28.8 & - & 75.5 & 70.3 \\
GCPM-G (Ours) & - & 42.0 & 0.8 & - & 45.0 & 2.1 & - & 47.0 & 30.9 \\
+MST Traversal & - & 53.7 & 47.5 & - & 57.6 & 23.2 & - & 76.2 & 72.1 \\
Max $\Delta\uparrow$ & - & \color{NavyBlue}{\textbf{17.9}$\uparrow$} & \color{NavyBlue}{\textbf{55.7}$\uparrow$} & - & \color{NavyBlue}{\textbf{15.5}$\uparrow$} & \color{NavyBlue}{\textbf{28.4}$\uparrow$} & - & \color{NavyBlue}{\textbf{34.2}$\uparrow$} & \color{NavyBlue}{\textbf{72.1}$\uparrow$} \\
\bottomrule
\end{tabular}
}
\label{tab: gc_results}
\vspace{-0.3cm}
\end{table}

\paragraph{Baselines and Clustering Algorithms}
We compare with state-of-the-art baselines in 
each task: \cite{grill2020bootstrap, li2021contrastive, yu2023contextually, wagstaff2001constrained, wu2019deep, li2023dual, metaxas2023divclust, zhong2021graph, liu2024interactive, tao2021clustering, ji2019invariant, lilearning, tsai2020mice, dang2021nearest, li2020prototypical, huang2020deep, huang2022learning, van2020scan, niu2022spice, shen2021you, li2022twin, sohn2020fixmatch, qian2023stable, kwon2023image} in GC; \cite{hu2017finding, guerin2018improving, ren2022diversified, yao2023augdmc, yao2024dual, yao2024multi, yao2024customized, liu2026conditional} in MC; and \cite{ji2019invariant, chen2020simple, he2020momentum, van2020scan, qian2023stable, chen2016infogan, singh2019finegan, li2020mixnmatch, benny2020onegan, rombach2022high, kim2021contrastive, yang2024dific, wang2023bridge} in FC. 
For LC, since each sample contains an image and an item name in the ABO dataset, we compare IC$|$TC~\cite{kwon2023image} using the same MLLM as the baseline. 
Due to the space limitation, we discuss details and our difference from baselines in Appendix~\ref{sec: compare_difference}. Note that many baselines require training and assume known numbers of clusters. For clustering algorithms, we select K-Means and HDBSCAN upon GCPM depending on whether the number of clusters is known in advance. Additionally, we test MST-based LLM Traversal upon HDBSCAN to validate its effectiveness for improving clustering results.
Note that for automatic cluster discovery, ACC is omitted because the number of predicted clusters may be different from the ground truths. Experiments are repeated with three random seeds and averaged.

\begin{table*}[t]
\centering
\caption{Performance comparison on multiple clustering, criteria, and metrics with averaged results per dataset. Baselines compared all assume known number of clusters. KMS.: K-Means; HDBS.: HDBSCAN. Max $\Delta\uparrow$ measures improvements of MST Traversal.}
\vspace{-0.2cm}
\scalebox{0.7}{
\begin{tabular}{l
  cc|cc|cc
  |cc|cc|cc
  |cc|cc|cc}
\toprule
\multirow{4}{*}{Method} 
& \multicolumn{6}{c|}{Fruit} 
& \multicolumn{6}{c|}{Card} 
& \multicolumn{6}{c}{CIFAR10-MC} \\
\cmidrule(lr){2-7} \cmidrule(lr){8-13} \cmidrule(lr){14-19}
& \multicolumn{2}{c|}{Color} & \multicolumn{2}{c|}{Species} & \multicolumn{2}{c|}{Average} 
& \multicolumn{2}{c|}{Number} & \multicolumn{2}{c|}{Suits} & \multicolumn{2}{c|}{Average} 
& \multicolumn{2}{c|}{Type} & \multicolumn{2}{c|}{Environment} & \multicolumn{2}{c}{Average} \\
\cmidrule(lr){2-3} \cmidrule(lr){4-5} \cmidrule(lr){6-7}
\cmidrule(lr){8-9} \cmidrule(lr){10-11} \cmidrule(lr){12-13}
\cmidrule(lr){14-15} \cmidrule(lr){16-17} \cmidrule(lr){18-19}
& NMI & RI & NMI & RI & NMI & RI 
& NMI & RI & NMI & RI & NMI & RI 
& NMI & RI & NMI & RI & NMI & RI \\
\midrule
MSC~\cite{hu2017finding}        & 68.9 & 80.5 & 16.3 & 60.5 & 42.6 & 70.5 & 8.1 & 78.1 & 5.0 & 35.9 & 6.6 & 57.0 & 15.5 & 33.0 & 11.4 & 30.8 & 13.5 & 31.9 \\
MCV~\cite{guerin2018improving}        & 62.7 & 76.9 & 27.3 & 66.0 & 45.0 & 71.5 & 7.9 & 71.3 & 4.3 & 36.4 & 6.1 & 53.9 & 16.2 & 33.1 & 13.8 & 33.4 & 15.0 & 33.3 \\
ENRC~\cite{miklautz2020deep}       & 71.0 & 85.1 & 31.9 & 65.4 & 51.5 & 75.3 & 12.3 & 73.1 & 6.8 & 38.0 & 9.6 & 55.6 & 18.3 & 34.7 & 18.9 & 36.0 & 18.6 & 35.4 \\
iMClusts~\cite{ren2022diversified}   & 73.5 & 86.3 & 30.3 & 67.4 & 51.9 & 76.9 & 11.4 & 76.6 & 7.2 & 37.2 & 9.3 & 56.9 & 20.4 & 37.0 & 19.2 & 36.6 & 19.8 & 36.8 \\
AugDMC~\cite{yao2023augdmc}     & 85.2 & 91.1 & 35.5 & 74.0 & 60.4 & 82.6 & 14.4 & 82.7 & 8.7 & 42.3 & 11.6 & 62.5 & 28.6 & 45.2 & 29.3 & 46.9 & 29.0 & 46.1 \\
DDMC~\cite{yao2024dual}       & 89.7 & 93.8 & 37.6 & 76.2 & 63.7 & 85.0 & 15.6 & 83.3 & 9.3 & 64.7 & 12.5 & 74.0 & 39.9 & 58.3 & 37.8 & 55.5 & 38.9 & 56.9 \\
Multi-MaP~\cite{yao2024multi}  & 86.2 & 95.3 & \textbf{100.0} & \textbf{100.0} & 93.1 & 97.6 & 36.5 & 85.9 & 27.3 & 70.4 & 31.9 & 78.2 & 49.7 & 71.0 & 46.0 & 67.4 & 47.9 & 69.2 \\
CRL~\cite{liu2026conditional} & - & - & - & - & - & - & 46.6 & - & 60.9 & - & 53.8 & - & - & - & - & - & - & - \\
Multi-Sub~\cite{yao2024customized}  & 96.9 & 99.6 & \textbf{100.0} & \textbf{100.0} & 98.5 & 99.8 & 39.2 & 88.4 & 31.0 & 79.4 & 35.1 & 83.9 & 52.7 & 73.9 & 48.3 & 71.0 & 50.5 & 72.5 \\
\midrule
\multicolumn{3}{l|}{\textit{KMS. w/ Known \# of Clus.}}  &  &  &    &   &  &    &   &  &    &  &  &    &   &  &    &  \\
GCPM-I (Ours) & 97.4 & 98.6 & \textbf{100.0} & \textbf{100.0} & 98.7 & 99.3 & 35.3 & 86.7 & 57.3 & 97.0 & 46.3 & 91.8 & 50.7 & 70.9 & 47.9 & 69.5 & 49.3 & 70.2 \\
GCPM-E (Ours) & 98.8 & \textbf{99.9} & \textbf{100.0} & \textbf{100.0} & 99.4 & \textbf{100.0} & \textbf{91.1} & \textbf{99.1} & \textbf{89.0} & \textbf{97.2} & \textbf{90.0} & \textbf{98.2} & 51.6 & 71.3 & 48.3 & 69.5 & 50.0 & 70.4 \\
GCPM-G (Ours) & \textbf{99.8} & \textbf{100.0} & \textbf{100.0} & \textbf{100.0} & \textbf{99.9} & \textbf{100.0} & 82.0 & 95.8 & 30.6 & 74.9 & 56.3 & 85.4 & \textbf{55.2} & \textbf{75.1} & \textbf{50.1} & \textbf{71.4} & \textbf{52.6} & \textbf{73.2} \\
\midrule
\multicolumn{3}{l|}{\makecell[l]{\textit{HDBS. \& Our MST Traversal} \\ \textit{ w/o Known \# of Clus.}}}  &  &  &    &   &  &    &   &  &    &  &  &    &   &  &    &  \\
GCPM-I (Ours) & 41.9 & 67.5 & 47.2 & 68.9 & 44.6 & 68.2 & 46.3 & 92.3 & 28.2 & 75.0 & 37.3 & 83.7 & 44.6 & 69.0 & 40.9 & 60.1 & 42.8 & 64.6 \\
+MST Traversal & 94.5 & 96.7 & 96.7 & 98.8 & 95.6 & 97.8 & 61.3 & 93.5 & 58.5 & 85.1 & 59.9 & 89.3 & 47.1 & 70.2 & 42.8 & 62.1 & 45.0 & 66.2 \\
GCPM-E (Ours) & 48.2 & 69.7 & 48.9 & 70.3 & 48.6 & 70.0 & 47.4 & 92.3 & 28.7 & 75.0 & 38.1 & 83.7 & 44.5 & 69.6 & 41.6 & 60.7 & 43.1 & 65.2 \\
+MST Traversal & 97.6 & 97.3 & \textbf{100.0} & \textbf{100.0} & 98.8 & 98.6 & 80.3 & 96.7 & 70.5 & 89.8 & 75.4 & 93.2 & 46.9 & 70.7 & 43.7 & 62.5 & 45.3 & 66.6 \\
GCPM-G (Ours) & 46.2 & 68.4 & 47.1 & 68.8 & 46.7 & 68.6 & 50.3 & 92.5 & 29.9 & 75.2 & 40.1 & 83.9 & 46.0 & 71.1 & 42.8 & 62.8 & 44.4 & 67.0 \\
+MST Traversal & 95.3 & 97.2 & \textbf{100.0} & \textbf{100.0} & 97.7 & 98.6 & 72.1 & 95.1 & 36.9 & 75.9 & 54.5 & 85.5 & 48.6 & 73.8 & 44.2 & 63.7 & 46.4 & 68.8 \\
Max $\Delta\uparrow$ & \color{NavyBlue}{\textbf{52.6}$\uparrow$} & \color{NavyBlue}{\textbf{29.2}$\uparrow$} & \color{NavyBlue}{\textbf{52.9}$\uparrow$}  & \color{NavyBlue}{\textbf{31.2}$\uparrow$} & \color{NavyBlue}{\textbf{52.8}$\uparrow$} &\color{NavyBlue}{\textbf{30.2}$\uparrow$}  & \color{NavyBlue}{\textbf{32.9}$\uparrow$} & \color{NavyBlue}{\textbf{4.4}$\uparrow$} & \color{NavyBlue}{\textbf{41.8}$\uparrow$} & \color{NavyBlue}{\textbf{14.8}$\uparrow$} & \color{NavyBlue}{\textbf{37.4}$\uparrow$} & \color{NavyBlue}{\textbf{9.6}$\uparrow$} & \color{NavyBlue}{\textbf{2.6}$\uparrow$} & \color{NavyBlue}{\textbf{2.7}$\uparrow$} & \color{NavyBlue}{\textbf{2.1}$\uparrow$} & \color{NavyBlue}{\textbf{2.0}$\uparrow$} & \color{NavyBlue}{\textbf{2.4}$\uparrow$} & \color{NavyBlue}{\textbf{2.4}$\uparrow$} \\
\bottomrule
\end{tabular}
}
\label{tab: mc_results}
\vspace{-0.5cm}
\end{table*}

\subsection{Experimental Results}
\paragraph{Overall Performance} 
Tables \ref{tab: gc_results}, \ref{tab: mc_results}, \ref{tab: fc_results}, and \ref{tab: lc_results} present comprehensive experimental results across general clustering (GC), multiple clustering (MC), fine-grained clustering (FC), and long-tail clustering (LC) tasks. Our training-free framework consistently outperforms existing state-of-the-art methods that require dataset-specific training. Notably, in GC tasks (Table \ref{tab: gc_results}), GCPM-G with K-Means achieves 98.8$\%$ accuracy on ImageNet-10, surpassing the previous best training-based method IDCTCL by 1.6$\%$. Similarly, in MC tasks (Table \ref{tab: mc_results}), our method achieves 99.9$\%$ NMI on Fruit dataset, significantly outperforming Multi-Sub (98.5$\%$). These results demonstrate that our guideline-driven approach, leveraging the semantic understanding of modern vision-language models, can achieve superior performance without requiring task-specific training.

\paragraph{Impact of Embedding Models}
We validate our framework's robustness using three embedding models with varying architectures and capacities. As shown across Tables \ref{tab: gc_results}-\ref{tab: lc_results}, there is a clear performance hierarchy: MLLM-based embeddings (GCPM-G) generally outperform LLM-based embeddings (GCPM-E), which in turn surpass regular instruction-aware encoders (GCPM-I). For instance, on STL-10, GCPM-G achieves 98.8$\%$ accuracy compared to 97.9$\%$ for GCPM-E and 98.2$\%$ for GCPM-I. This trend reflects the enhanced multimodal reasoning capabilities of larger, more sophisticated models.

\begin{figure}
    \centering
    \includegraphics[width=1.0\linewidth]{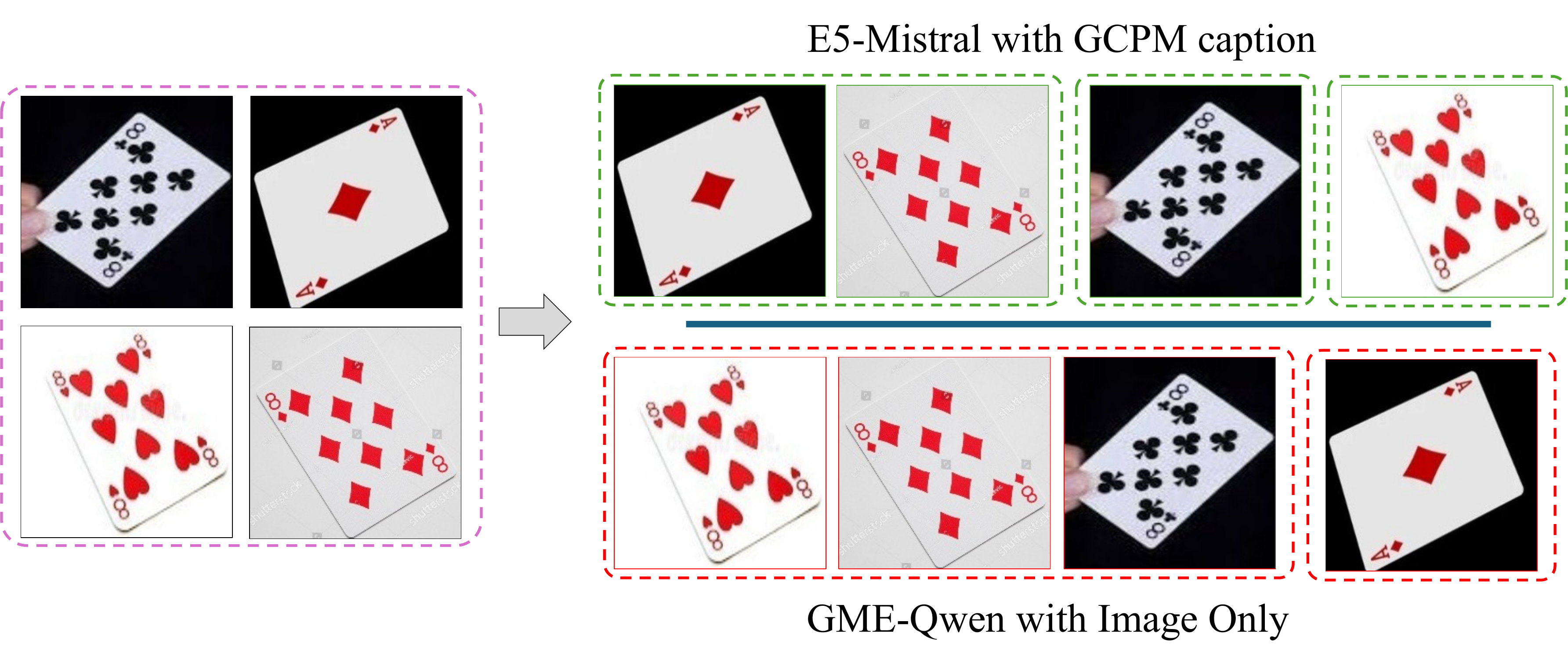}
    \vspace{-0.7cm}
    \caption{A disentanglement test case of cards \textit{grouped by suits} using HDBSCAN. While using GME-Qwen with images only, the number criteria dominates the suit criteria because of the layout.}
    \label{fig: disentanglement_example}
\vspace{-0.6cm}
\end{figure}

However, the Card dataset presents an interesting exception where GCPM-E (91.1$\%$ NMI for number criterion) outperforms GCPM-G (82.0$\%$ NMI). This counterintuitive result can be attributed to the dataset's inherent visual complexity because card numbers and suits are heavily visually entangled within single images. In such scenarios, the concept proxy approach of extracting textual descriptions before encoding proves more effective than direct multimodal embedding, as it explicitly disentangles these tangled visual attributes through the captioning process (also see a test case in Figure~\ref{fig: disentanglement_example}). This finding highlights our framework's adaptability: the GCPM effectively handles scenarios where visual features are difficult to differentiate.

\paragraph{Clustering Algorithm Selection}
Comparing K-Means and HDBSCAN reveals intricate trade-offs. When cluster numbers are known, K-Means consistently achieves superior performance across most datasets, aligning with existing findings~\cite{li2023image}. For example, on CIFAR-10, GCPM-G with K-Means achieves 94.1$\%$ accuracy compared to 42.0$\%$ NMI with HDBSCAN alone. However, this advantage diminishes in highly imbalanced scenarios. In the ABO-LC dataset (Table \ref{tab: lc_results}), where 78.7$\%$ of clusters contain two or fewer samples, HDBSCAN with MST Traversal achieves better performance (51.5$\%$ ARI) without requiring prior knowledge of cluster counts, demonstrating its suitability for real-world long-tail distributions.

\paragraph{Effectiveness of MST Traversal}
The MST-based LLM Traversal consistently improves upon HDBSCAN's initial clustering across all experimental settings. The magnitude of improvement varies systematically across task types. In GC tasks, improvements are substantial. For instance, on ImageNet-10, MST Traversal improves ARI from 0.3 to 72.1. In MC tasks, improvements are more variable across different criteria: significant gains are observed for criteria requiring abstract semantic understanding (e.g., 52.9$\uparrow$ in NMI for Fruit species), while criteria with clear visual patterns show modest gains.

Interestingly, improvements in FC tasks (Table \ref{tab: fc_results}) are relatively modest compared with in GC tasks.
This behavior reflects the fundamental difference of fine-grained clustering from general clustering: distinguishing subtle discrepancy between similar categories demands extremely precise semantic judgments. 
Therefore, we use conservative prompts for the LLM merging process in fine-grained clustering. These prompts maintain strict rules while comparing two clusters and keep them separated unless there are clear difference in specific criteria required by guidelines.
This conservative behavior, however, still yields state-of-the-art results, demonstrating that the combination of high-quality guideline-aware embeddings and selective LLM refinement is effective even in challenging fine-grained scenarios.

\begin{table}[!t]
\centering
\caption{Comparison on fine-grained clustering. Baselines except UFCL assume known cluster number. KMS.: K-Means; HDBS.: HDBSCAN. Max $\Delta\uparrow$ measures improvements of MST Traversal.}
\vspace{-0.2cm}
\setlength{\tabcolsep}{4pt}
\scalebox{0.7}{
\begin{tabular}{lcc|cc|cc|cc}
\toprule
\multirow{2}{*}{Method} 
& \multicolumn{2}{c|}{CUB Birds} 
& \multicolumn{2}{c|}{Stanford Dogs} 
& \multicolumn{2}{c|}{Stanford Cars} 
& \multicolumn{2}{c}{Oxford Flowers} \\
\cmidrule(lr){2-3} \cmidrule(lr){4-5} \cmidrule(lr){6-7} \cmidrule(lr){8-9}
& ACC & NMI & ACC & NMI & ACC & NMI & ACC & NMI \\
\midrule
IIC~\cite{ji2019invariant}              & 7.4  & 36.0 & 5.0  & 18.0 & 4.9  & 27.0 & 8.7  & 24.0 \\
SimCLR~\cite{chen2020simple}           & 8.4  & 40.0 & 6.8  & 19.0 & 6.7  & 33.0 & 12.5 & 29.0 \\
MoCo~\cite{he2020momentum}             & 10.2 & 37.0 & 11.5 & 31.0 & 8.0  & 37.0 & 51.5 & 68.0 \\
SCAN~\cite{van2020scan}             & 11.9 & 45.0 & 12.3 & 35.0 & 8.8  & 38.0 & 56.5 & 77.0 \\
SeCu~\cite{qian2023stable}             & 15.4 & 48.0 & 16.7 & 38.0 & 10.0 & 39.0 & 68.6 & 84.0 \\
InfoGAN~\cite{chen2016infogan}          & 8.6  & 39.0 & 6.4  & 21.0 & 6.5  & 31.0 & 23.2 & 44.0 \\
FineGAN~\cite{singh2019finegan}          & 12.6 & 40.0 & 7.9  & 23.0 & 7.8  & 35.0 & -    & -    \\
MixNMatch~\cite{li2020mixnmatch}        & 13.6 & 42.0 & 8.9  & 32.0 & 7.9  & 36.0 & -    & -    \\
OneGAN~\cite{benny2020onegan}           & 10.1 & 39.0 & 7.3  & 21.0 & 6.0  & 27.0 & -    & -    \\
SD~\cite{rombach2022high} & 7.6  & 34.0 & 5.4  & 18.0 & 9.1  & 37.0 & 13.1 & 34.0 \\
C3-GAN~\cite{kim2021contrastive}           & 22.7 & 50.0 & 11.8 & 30.0 & 8.3  & 33.0 & 55.6 & 72.0 \\
DiFiC~\cite{yang2024dific}            & 31.7 & 61.0 & 17.2 & 39.0 & 47.2 & 68.0 & 72.9 & 88.0 \\
UFCL~\cite{wang2023bridge} & - & 78.6 & - & 45.8 & - & 46.5 & - & 70.7 \\
\midrule
\multicolumn{3}{l|}{\textit{KMS. w/ Known \# of Clus.}}  &  &  &    &  &  \\
GCPM-I (Ours) & 41.2 & 70.3 & 54.5 & 68.8 & 69.4 & 87.7 & 73.9 & 89.0 \\
GCPM-E (Ours) & 45.5 & 74.1 & 59.8 & 74.0 & \textbf{74.6} & \textbf{90.0} & 77.4 & 90.6 \\
GCPM-G (Ours) & \textbf{72.9} & \textbf{89.9} & \textbf{75.1} & \textbf{85.9} & 66.3 & 86.2 & \textbf{86.0} & \textbf{94.9} \\
\midrule
\multicolumn{3}{l|}{\makecell[l]{\textit{HDBS. \& Our MST Traversal} \\ \textit{ w/o Known \# of Clus.}}}&  &  &  &  &  &  \\
GCPM-I (Ours) & - & 75.7 & - & 70.1 & - & 78.3 & - & 75.9 \\
+MST Traversal & - & 76.3 & - & 69.7 & - & 81.2 & - & 77.5 \\
GCPM-E (Ours) & - & 77.9 & - & 71.4 & - & 80.2 & -  & 82.7 \\
+MST Traversal & - & 78.4 & - & 72.1 & - & 83.4 & - & 84.2 \\
GCPM-G (Ours) & - & 84.9 & - & 76.5 & - & 78.8 & - & 88.2 \\
+MST Traversal & - & 85.9 & - & 78.0 & - & 80.9 & -  & 88.6 \\
Max $\Delta\uparrow$ & - & \color{NavyBlue}{\textbf{1.0}$\uparrow$} & - & \color{NavyBlue}{\textbf{1.5}$\uparrow$} & - & \color{NavyBlue}{\textbf{3.2}$\uparrow$} & - & \color{NavyBlue}{\textbf{1.6}$\uparrow$} \\ 
\bottomrule
\end{tabular}
}
\label{tab: fc_results}
\vspace{-0.4cm}
\end{table}

\subsection{In-Depth Analysis}
\paragraph{How MST Traversal Affects Clustering Results?}
To further analyze the effectiveness of MST Traversal, we employ multi-faceted metrics analogous to those used in classification tasks, rather than relying solely on general metrics such as NMI. Specifically, we adopt BCubed~\cite{bagga1998entity} Precision and Recall metrics. Results are presented in Table \ref{tab: bcubed_results}. We observe that HDBSCAN alone typically produces high precision but very low recall. This indicates that the model successfully groups obviously similar samples into small clusters but fails to form larger, more comprehensive clusters. This limitation likely stems from the difficulty of aggregating semantically complex samples using embeddings alone.

\begin{table}[!t]
\centering
\caption{Comparison on ABO-LC. Baseline IC$|$TC is based on known number of clusters. KMS.: K-Means; HDBS.: HDBSCAN. Max $\Delta\uparrow$ measures improvements of MST Traversal.}
\vspace{-0.2cm}
\setlength{\tabcolsep}{8pt}
\scalebox{0.75}{
\begin{tabular}{l|c|c|c}
\toprule
& ACC & NMI & ARI \\
\midrule
IC$|$TC~\cite{kwon2023image}  & 5.5 & 35.3 & 5.3\\
\midrule
\textit{KMS. w/ Known \# of Clus.}  &  &  & \\
GCPM-I (Ours)  & \textbf{55.7} & 92.9 & 38.4 \\
GCPM-E (Ours)  & 54.9 & 92.7 & 36.2 \\
GCPM-G (Ours)  & 55.2 & 93.0 & 32.3 \\
\midrule
\makecell[l]{\textit{HDBS. \& Our MST Traversal} \\ \textit{ w/o Known \# of Clus.}}  &  &  &  \\
GCPM-I (Ours)  & - & 92.4 & 27.5 \\
+MST Traversal  & - & \textbf{93.3} & 50.7 \\
GCPM-E (Ours)  & - & 92.3 & 28.2 \\
+MST Traversal  & - & 93.1 & \textbf{51.5} \\
GCPM-G (Ours)  & - & 92.1 & 23.5 \\
+MST Traversal  & - & 92.9 & 37.9 \\
Max $\Delta\uparrow$ & - & \color{NavyBlue}{\textbf{0.9}$\uparrow$}  & \color{NavyBlue}{\textbf{23.3}$\uparrow$}  \\
\bottomrule
\end{tabular}
}
\label{tab: lc_results}
\vspace{-0.5cm}
\end{table}

In contrast, K-Means produces comparable precision and recall. This balanced performance can be attributed to the prior knowledge of cluster numbers, which forces the algorithm to aggregate more samples into larger groups. After applying MST Traversal, we observe a substantial improvement in Recall with only a modest sacrifice in Precision. It demonstrates that the LLM successfully merges most small clusters into larger, semantically coherent groups according to the guideline. The slight decrease in Precision can be attributed to two factors: (1) guidelines are generated in an unsupervised manner without explicit ground truth, potentially introducing ambiguity, and (2) the LLM may occasionally struggle with interpreting subtle guideline differences. We discuss strategies~\cite{Zhong_Li_Dang_Jiang_Ma_Guo_Gao_Huang_2026} for improving guideline precision through prompt optimization in Appendix~\ref{sec: improve_guideline}.

\paragraph{Effectiveness of GCPM}
While the text-only embeddings have already validated GCPM's effectiveness, we conduct additional experiments comparing it against using images alone or using standard captioning prompts from \cite{liu2023visual} instead of guideline. Table \ref{tab: GCPM_MST_ablations} shows that when using GME-Qwen as the embedding model, incorporating concept proxy captions consistently improves performance across all datasets. This improvement demonstrates that GCPM's guideline-aware captioning process effectively surfaces relevant attributes that might otherwise be overshadowed in direct image encoding, validating our design choice to employ concept proxy extraction as an intermediate step.

\begin{figure}[!t]
    \centering
    \includegraphics[width=1.0\linewidth]{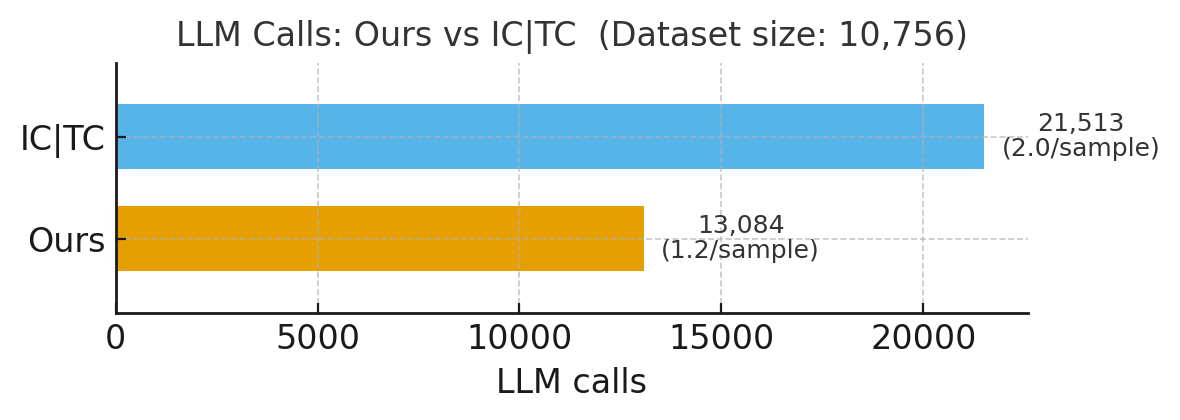}
    \vspace{-0.7cm}
    \caption{Comparison of LLM calls used for clustering and their ratio to the number of samples on the ABO-LC dataset.}
    \label{fig: llm_calls}
\vspace{-0.5cm}
\end{figure}

\paragraph{Runtime Analysis of MST Traversal}
To eliminate the confounding effects of hardware variations and different LLM implementations, we measure runtime using the number of LLM calls as our primary metric. Results are in Table \ref{tab: llm_calls} and Figure~\ref{fig: llm_calls}. Compared to naive pairwise matching where the LLM directly evaluates all cluster pairs, our MST Traversal demonstrates significantly higher efficiency. Notably, MST Traversal is initialized after HDBSCAN, meaning it operates on clusters rather than individual samples, which substantially reduces the number of merge operations required compared with \cite{kwon2023image, chen2025agentcentricpersonalizedmultipleclustering}. Furthermore, we cache LLM inference results after each decision. When identical cluster pairs appear in subsequent iterations, we directly utilize cached results, thereby avoiding redundant LLM calls and further improving computational efficiency.


\paragraph{More Ablations}
We discuss more ablation study, including hyperparameter analysis, GCPM with other clustering algorithms, MST Traversal with proprietary LLMs, and using MST Traversal for existing baselines in Appendix~\ref{sec: more_ablations}.

\begin{table}[t]
\centering
\caption{Comparison of clustering results based on BCubed Precision (B-Prec.) and Recall (B-Rec.) before and after using MST Traversal upon HDBSCAN. \# of clusters includes singletons.}
\vspace{-0.2cm}
\setlength{\tabcolsep}{7pt}
\scalebox{0.7}{
\begin{tabular}{l
  |ccc
  |ccc}
\toprule
& \multicolumn{3}{c|}{ImageNet-10} 
& \multicolumn{3}{c}{Card-Number} \\
\cmidrule(lr){2-4} \cmidrule(lr){5-7}
& \# of Clus. & B-Prec. & B-Rec. 
& \# of Clus. & B-Prec. & B-Rec. \\
\midrule
K-Means & 10 & 98.6 & 98.6 & 13 & 81.1 & 81.4 \\
\midrule
Before & 7034 & 99.7 & 19.9 & 4191 & 98.6 & 3.1 \\
After  & 251 & 93.5 & 62.3 & 151 & 90.6 & 42.3 \\
\bottomrule
\end{tabular}
}
\label{tab: bcubed_results}
\vspace{-0.4cm}
\end{table}
\begin{table}[t]
\centering
\caption{Comparison of caption quality across datasets using different captioning strategies via K-Means.}
\vspace{-0.2cm}
\setlength{\tabcolsep}{4pt}
\scalebox{0.7}{
\begin{tabular}{l|c|c|c}
\toprule
& ImageNet-10 & Card-Number & Stanford Cars \\
\midrule
GME-QWen w/ Image Only          & 94.7 & 71.9 & 61.5 \\
GME-QWen w/ Standard Caption  & 93.7 & 73.3 & 69.2 \\
GME-QWen w/ GCPM Caption      & 96.7 & 82.0 & 86.2 \\
\bottomrule
\end{tabular}
}
\label{tab: GCPM_MST_ablations}
\vspace{-0.4cm}
\end{table}
\begin{table}[t]
\centering
\caption{Comparison of number of LLM calls in MST Traversal across datasets and their ratios to the number of samples.}
\vspace{-0.2cm}
\setlength{\tabcolsep}{5.5pt}
\scalebox{0.75}{
\begin{tabular}{l|c|c|c}
\toprule
& ImageNet-10 & Card-Number & Stanford Cars \\
\# of Samples & (13000) & (8029) & (8041) \\
\midrule
MST Traversal & 11232 & 6506 & 10803 \\
\midrule
LLM / Sample Ratio & 0.86 & 0.81 & 1.34 \\
\bottomrule
\end{tabular}
}
\label{tab: llm_calls}
\vspace{-0.4cm}
\end{table}


\section{Conclusion}
We introduce a universal clustering framework that bridges fundamental gaps in diverse clustering scenarios through textual guidelines. Our hybrid approach combines Generative Concept Proxy Modeling for efficient guideline-aware embeddings with MST-based LLM Traversal for selective semantic reasoning, achieving superior performance without task-specific training. Comprehensive experiments across general, multiple, fine-grained, and long-tail clustering tasks demonstrate the superiority of our method. 

\section{Acknowledgement}
This work was partially supported by US National Science Foundation IIS-2412195, CCF-2400785, the Cancer Prevention and Research Institute of Texas (CPRIT) award (RP230363), the National Institutes of Health (NIH) R01 award (1R01AI190103-01) and Microsoft Accelerate Foundation Models Research (2024).

%% file: sec/X_suppl.tex
\clearpage
\setcounter{page}{1}
\maketitlesupplementary

\renewcommand{\thesection}{\Alph{section}}
\setcounter{section}{0}

\section*{Appendix Table of Contents}
\newcommand{\apptoc}[2]{%
  \noindent\makebox[\linewidth][l]{#1\dotfill\ #2}\par
}
\apptoc{\textbf{A. Computational Complexity}}{p.~1}
\apptoc{\hspace*{1.5em}A.1. Proof of Computational Complexity}{p.~1}
\apptoc{\hspace*{3.0em}A.1.1. Problem Setup}{p.~1}
\apptoc{\hspace*{3.0em}A.1.2. Probabilistic Model for LLM Decisions}{p.~1}
\apptoc{\hspace*{3.0em}A.1.3. Case 1: High Merge Rate Regime}{p.~1}
\apptoc{\hspace*{3.0em}A.1.4. Case 2: Adaptive Merge Rate Model}{p.~2}
\apptoc{\hspace*{3.0em}A.1.5. Total LLM Calls Calculation}{p.~2}
\apptoc{\hspace*{3.0em}A.1.6. Main Results}{p.~2}
\apptoc{\hspace*{1.5em}A.2. In-Depth Runtime Complexity Analysis}{p.~2}
\vspace{0.5\baselineskip}
\apptoc{\textbf{B. Dataset Details}}{p.~3}
\apptoc{\hspace*{1.5em}B.1. ABO Data Processing}{p.~3}
\apptoc{\hspace*{1.5em}B.2. Existing Datasets and Metrics}{p.~3}
\vspace{0.5\baselineskip}
\apptoc{\textbf{C. Guideline Processing}}{p.~5}
\apptoc{\hspace*{1.5em}C.1. Guideline Generation}{p.~5}
\apptoc{\hspace*{1.5em}C.2. Detailed Guideline}{p.~5}
\apptoc{\hspace*{1.5em}C.3. Guideline and Image Forwarding Process}{p.~5}
\apptoc{\hspace*{1.5em}C.4. Improving Guideline Accuracy}{p.~7}
\vspace{0.5\baselineskip}
\apptoc{\textbf{D. Comparison with Existing Methods}}{p.~7}
\apptoc{\hspace*{1.5em}D.1. Guideline-aware and Text-guided Clustering}{p.~7}
\apptoc{\hspace*{1.5em}D.2. LLM-based Clustering}{p.~8}
\apptoc{\hspace*{1.5em}D.3. General Clustering Baselines}{p.~9}
\apptoc{\hspace*{1.5em}D.4. Multiple Clustering Baselines}{p.~9}
\apptoc{\hspace*{1.5em}D.5. Fine-grained Clustering Baselines}{p.~10}
\vspace{0.5\baselineskip}
\apptoc{\textbf{E. More Ablations}}{p.~10}
\apptoc{\hspace*{1.5em}E.1. Hyperparameter Analysis}{p.~10}
\apptoc{\hspace*{1.5em}E.2. GCPM with Other Clustering Algorithms}{p.~10}
\apptoc{\hspace*{1.5em}E.3. MST Traversal with Proprietary LLMs}{p.~11}
\apptoc{\hspace*{1.5em}E.4. Using MST Traversal for Existing Baselines}{p.~11}
\vspace{0.5\baselineskip}
\apptoc{\textbf{F. Reproduction and Limitations}}{p.~11}

\section{Computational Complexity}
\subsection{Proof of Computational Complexity}
\label{sec: proof_complexity}
In this section, we analyze the computational complexity of LLM calls with probabilistic models. We first define notations and assumptions used for analysis. Then, we discuss a simplified scenario for complexity and further extend it to a more realistic scenario.
\subsubsection{Problem Setup}
\begin{definition}
Let us define the mathematical framework for the clustering algorithm:
\begin{align}
k_i &= \text{number of clusters at the start of iteration } i \\
m_i &= \text{number of merges performed in iteration } i \\
c_i &= \text{number of LLM calls in iteration } i = k_i - 1
\end{align}
\end{definition}
\textbf{Constraints:}
\begin{align}
&k_1 = N \quad \text{(initial condition)} \\
&k_{i+1} = k_i - m_i \quad \text{(recurrence relation)} \\
&m_i \leq \lfloor k_i/2 \rfloor \quad \text{(non-overlapping merge constraint)}
\end{align}
\textbf{Termination:} $\quad k_i = 1$ or $m_i = 0$

\subsubsection{Probabilistic Model for LLM Decisions}
\textbf{Assumption:} Each MST edge is independently approved by the LLM with probability $p$.

For $k$ clusters with MST having $k-1$ edges, the expected number of actual merges is:
\begin{equation}
\mathbb{E}[m_i] = \min(\alpha p(k_i-1), \lfloor k_i/2 \rfloor)
\end{equation}
where $\alpha \leq 1$ accounts for merge conflicts due to non-overlapping constraints.
\subsubsection{Case 1: High Merge Rate Regime}
When $\alpha p(k_i-1) \leq k_i/2$, we have:
\begin{align}
\mathbb{E}[k_{i+1}] &= k_i - \alpha p(k_i-1) \\
&= k_i(1-\alpha p) + \alpha p
\end{align}

For large $k_i$, this approximates to:
\begin{equation}
\mathbb{E}[k_{i+1}] \approx k_i(1-\alpha p)
\end{equation}

\textbf{Exponential Decay Analysis:}
\begin{equation}
k_i \approx N(1-\alpha p)^{i-1}
\end{equation}

\textbf{Number of iterations} until $k_T = 1$:
\begin{align}
N(1-\alpha p)^{T-1}&= 1 \\
T = \frac{\log(N)}{\log(1/(1-\alpha p))} + 1 &= \mathcal{O}(\log N)
\end{align}

\textbf{Total LLM calls:}
\begin{align}
\sum_{i=1}^T c_i &= \sum_{i=1}^T (k_i - 1) \\
&= \sum_{i=1}^T (N(1-\alpha p)^{i-1} - 1) \\
&= N \sum_{i=1}^T (1-\alpha p)^{i-1} - T \\
&= N \cdot \frac{1-(1-\alpha p)^T}{1-(1-\alpha p)} - T \\
&= \frac{N(1-(1-\alpha p)^T)}{\alpha p} - \mathcal{O}(\log N)
\end{align}

Since $(1-\alpha p)^T \approx 1/N$:
\begin{align}
    \sum_{i=1}^T c_i &\approx \frac{N(1-1/N)}{\alpha p} - \mathcal{O}(\log N)\\
    &= \frac{N}{\alpha p} - \mathcal{O}(\log N)\\
    &= \mathcal{O}(N)
\end{align}

\subsubsection{Case 2: Adaptive Merge Rate Model}
\textbf{More realistic assumption:} Merge probability decreases as clusters become more refined.

Let $p_i = \frac{c}{\log(k_i)}$ for constant $c > 0$.

Expected merges:
\begin{equation}
\mathbb{E}[m_i] \approx \alpha \cdot \frac{c}{\log(k_i)} \cdot (k_i-1) = \frac{\alpha c(k_i-1)}{\log(k_i)}
\end{equation}

For large $k_i$:
\begin{equation}
\mathbb{E}[k_{i+1}] \approx k_i - \frac{\alpha c k_i}{\log(k_i)} = k_i\left(1 - \frac{\alpha c}{\log(k_i)}\right)
\end{equation}

\textbf{Continuous approximation:}
\begin{equation}
\frac{dk}{dt} = -\frac{\alpha c k}{\log(k)}
\end{equation}

Separating variables:
\begin{equation}
\frac{\log(k)}{k} dk = -\alpha c \, dt
\end{equation}

Integrating both sides:
\begin{equation}
\int \frac{\log(k)}{k} dk = -\alpha c \int dt
\end{equation}

Let $u = \log(k)$, then $du = \frac{dk}{k}$:
\begin{equation}
\int u \, du = \frac{(\log(k))^2}{2}
\end{equation}

Therefore:
\begin{equation}
\frac{(\log(k))^2}{2} = -\alpha c t + C
\end{equation}

\textbf{Initial condition:} $k(0) = N$, so $C = \frac{(\log(N))^2}{2}$

\begin{equation}
(\log(k))^2 = (\log(N))^2 - 2\alpha c t
\end{equation}

\textbf{Termination time} when $k = 1$ (i.e., $\log(k) = 0$):
\begin{align}
0 &= (\log(N))^2 - 2\alpha c T \\
T &= \frac{(\log(N))^2}{2\alpha c} = \mathcal{O}((\log N)^2)
\end{align}

\subsubsection{Total LLM Calls Calculation}
\textbf{LLM calls per iteration:} $k_i - 1$

\textbf{Total calls:}
\begin{equation}
\sum_{i=1}^T (k_i - 1) \approx \int_0^T k(t) \, dt - T
\end{equation}

From our differential equation solution:
\begin{equation}
k(t) = \exp\left(\sqrt{(\log(N))^2 - 2\alpha c t}\right)
\end{equation}

Since $k(t)$ decreases from $N$ to $1$ over $T = \mathcal{O}((\log N)^2)$ iterations, and accounting for the non-uniform decay pattern, the integral evaluates to:
\begin{equation}
\text{Total calls} = \mathcal{O}(N \log N)
\end{equation}

\subsubsection{Main Results}
\begin{theorem}
Under the adaptive merge rate model $p_i = \frac{c}{\log(k_i)}$, the expected number of LLM calls for the hierarchical clustering algorithm is $\mathcal{O}(N \log N)$.
\end{theorem}

\subsection{In-Depth Runtime Complexity Analysis}
\label{sec: runtime_complexity}
In this section, we validate the assumption of the adaptive merge rate model from Section~\ref{sec: proof_complexity}. We then discuss how caching rejected pairs accelerates the MST Traversal.

\begin{figure}[!t]
    \centering
    \includegraphics[width=1.0\linewidth]{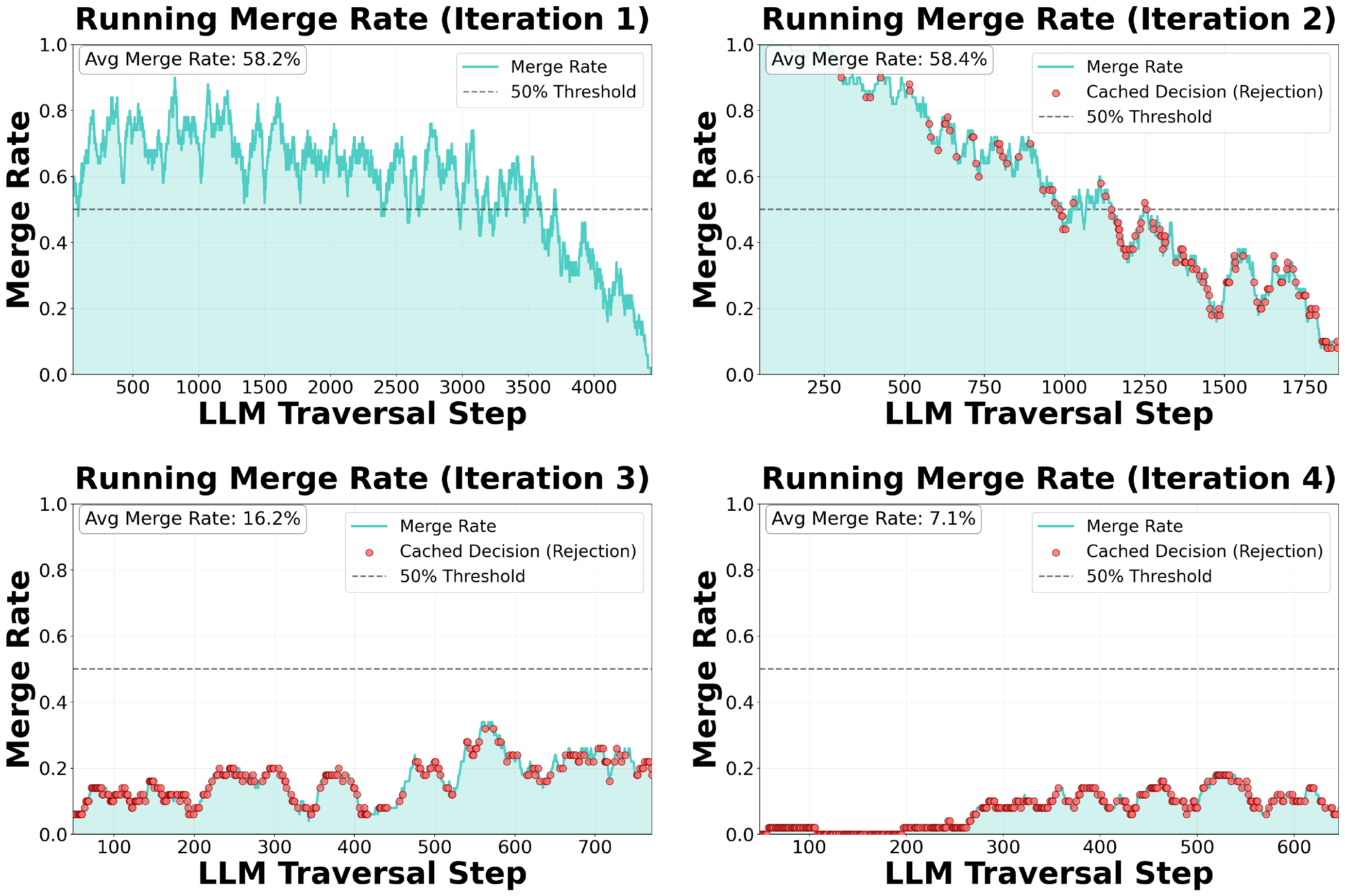}
    \caption{Merging rates over four iterations. The running merging rate is calculated by moving average over 50 consecutive steps. Red dots represents cached rejected decision that the LLM can safely skip. The merging rates decreases as iteration proceeds. The cached decisions significantly reduce the number of LLM calls.}
    \label{fig: merge_rate}
\end{figure}

We track merging patterns during the traversal process. As mentioned in Section 3.3, we build a traversal path based on the edge weights of the MST in ascending order. Each edge weight measures how close a pair of clusters is. The LLM makes a decision on whether they should be merged following this order. We track all merging decisions for MST Traversal across different iterations on the Card dataset conditioned on number. Results are in Figure~\ref{fig: merge_rate}.

We observe that merging occurs more frequently among earlier pairs and gradually decreases as the traversal proceeds. This validates the effectiveness of our hybrid design: guideline-driven embeddings provide a reliable distance matrix such that closer pairs have higher priority to be examined by the LLM. This also validates the adaptive merge rate model assumption that the LLM merges clusters more frequently at the beginning, and the merge probabilities decrease as clusters become more refined.
Notably, if the LLM rejects a pair, they will exist in the next iteration and may be selected again by the MST. As merging becomes less frequent in later iterations, more and more rejected pairs will be repeatedly selected. Therefore, we cache decisions for all rejected pairs and skip them if they occur subsequently. This cache mechanism reduces the number of LLM calls and thus improves efficiency.

\section{Dataset Details}
\begin{figure}[t!]
    \centering
    \includegraphics[width=1.0\linewidth]{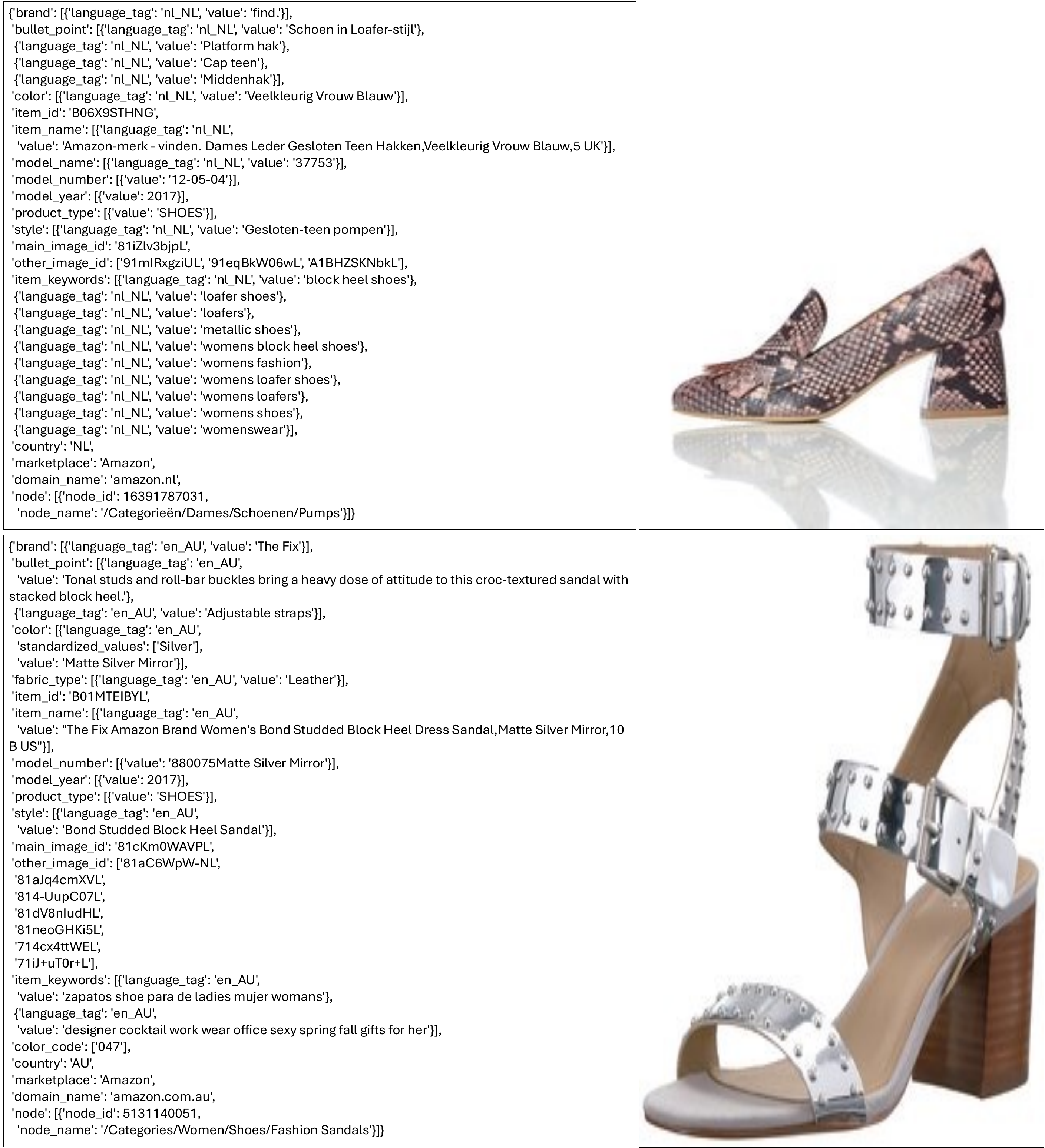}
    \caption{Raw sample examples from the original ABO dataset.}
    \label{fig: abo_raw_examples}
\end{figure}
\subsection{ABO Data Processing}
\label{sec: abo_processing}
We downloaded 398,212 original images and metadata from the ABO~\cite{collins2022abo} dataset official website\footnote{https://amazon-berkeley-objects.s3.amazonaws.com/index.html}. Each item contains an image and multiple textual attributes as shown in Figure~\ref{fig: abo_raw_examples}. We select several key attributes reflecting how items are typically categorized in modern e-commerce systems: ['brand', 'color', 'item\_id', 'model\_name', 'product\_type', 'style']. Subsequently, we filtered out items without these attributes, resulting in 10,756 items. We then apply the group\_by function from the pandas package\footnote{https://pandas.pydata.org/docs/index.html} to these items based on all key attributes. Each item is ultimately represented by its image and item name, which are not provided during the grouping process. The item name and image provide complementary textual (e.g., brand, model name) and visual (e.g., color, style) information to infer the key attributes of a given item. Note that unlike existing public datasets that are originally designed for classification, the ABO-LC dataset does not have an explicit definition for each category, reflecting real-world scenarios and making the clustering process even more challenging.

\subsection{Existing Datasets and Metrics}
\label{sec: data_and_metrics}
\paragraph{Metrics}
In general, we use Accuracy (ACC), Normalized Mutual Information (NMI), Rand Index (RI), Adjusted Rand Index (ARI), Bcubed Precision (B-Prec.) and Recall (B-Rec) in the paper. 
Given two cluster assignments 
$Y = \{y_1, \dots, y_N\}$ and 
$C = \{c_1, \dots, c_N\}$,
the Normalized Mutual Information (NMI) is defined as
\begin{equation}
\mathrm{NMI}(Y, C) = \frac{2\, I(Y; C)}{H(Y) + H(C)},
\end{equation}
where $I(Y; C)$ denotes the mutual information between $Y$ and $C$, and
\begin{equation}
I(Y; C) = \sum_{i=1}^{K} \sum_{j=1}^{L} 
p_{ij} \log \frac{p_{ij}}{p_i \, p_j},
\end{equation}
with $p_i$, $p_j$ the marginal probabilities and $p_{ij}$ the joint probability
of $Y=i$ and $C=j$. The entropies are defined as
\begin{equation}
H(Y) = - \sum_{i=1}^{K} p_i \log p_i,
\qquad
H(C) = - \sum_{j=1}^{L} p_j \log p_j .
\end{equation}
Given the ground-truth labels 
$Y = \{y_1, \dots, y_N\}$ 
and the predicted cluster assignments 
$C = \{c_1, \dots, c_N\}$,
define the contingency matrix
\[
n_{ij} = 
\bigl|\{\, k \mid y_k = i,\ c_k = j \,\}\bigr|.
\]
To compare clusters and true labels, we seek a permutation 
$\pi$ of the cluster indices that maximizes the matching:
\[
\max_{\pi} \sum_{k=1}^{N} 
\mathbf{1}[\, y_k = \pi(c_k) \,].
\]
The permutation $\pi$ is obtained by applying the Hungarian 
(Munkres) algorithm to the cost matrix derived from $-n_{ij}$.
After alignment, define the predicted labels 
$\hat{y}_k = \pi(c_k)$.
The clustering accuracy is then
\begin{equation}
\mathrm{ACC}(Y,C)
=
\frac{1}{N}
\sum_{k=1}^{N}
\mathbf{1}[\, y_k = \hat{y}_k \,].
\end{equation}
The Rand Index (RI) evaluates the agreement of pairwise decisions.  
Let $a$ be the number of pairs $(i,j)$ for which 
$y_i = y_j$ and $c_i = c_j$,  
and let $b$ be the number of pairs $(i,j)$ for which 
$y_i \neq y_j$ and $c_i \neq c_j$.  
The Rand Index is defined as
\begin{equation}
\mathrm{RI}(Y, C)
= \frac{a + b}{\binom{N}{2}} .
\end{equation}
Let $n_{ij}$ denote the number of samples that belong to 
ground-truth class $i$ and predicted cluster $j$.
Define the row sums $a_i = \sum_{j} n_{ij}$ 
and column sums $b_j = \sum_{i} n_{ij}$.
The Adjusted Rand Index (ARI) is defined as
\begin{equation}
\begin{aligned}
&\mathrm{ARI}(Y, C)= \\
&\frac{
\displaystyle
\sum_{i} \sum_{j} \binom{n_{ij}}{2}
\;-\;
\frac{
\left(\sum_i \binom{a_i}{2}\right)
\left(\sum_j \binom{b_j}{2}\right)
}{
\binom{N}{2}
}
}{
\displaystyle
\frac{1}{2}
\left[
\sum_i \binom{a_i}{2}
+
\sum_j \binom{b_j}{2}
\right]
\;-\;
\frac{
\left(\sum_i \binom{a_i}{2}\right)
\left(\sum_j \binom{b_j}{2}\right)
}{
\binom{N}{2}
}
}.
\end{aligned}
\end{equation}
For each sample $i$, define the pairwise correctness indicator
\[
\mathrm{correct}(i,j) =
\begin{cases}
1, & \text{if } (y_i = y_j)\ \text{and}\ (c_i = c_j),\\[3pt]
0, & \text{otherwise}.
\end{cases}
\]
Let $\mathrm{pred}(i,j) = 1$ if $c_i = c_j$ and 
$\mathrm{gt}(i,j) = 1$ if $y_i = y_j$.
The Bcubed Precision for sample $i$ is
\begin{equation}
P_i = 
\frac{
\sum_{j \neq i} \mathrm{correct}(i,j)
}{
\sum_{j \neq i} \mathrm{pred}(i,j)
}.
\end{equation}
The Bcubed Recall for sample $i$ is
\begin{equation}
R_i = 
\frac{
\sum_{j \neq i} \mathrm{correct}(i,j)
}{
\sum_{j \neq i} \mathrm{gt}(i,j)
}.
\end{equation}
The final Bcubed Precision (B-Prec.) and Recall (B-Rec.) are the averages:
\begin{equation}
\text{B-Prec}
= \frac{1}{N} \sum_{i=1}^{N} P_i,
\qquad
\text{B-Rec}
= \frac{1}{N} \sum_{i=1}^{N} R_i .
\end{equation}

\paragraph{Datasets}
We use ten public datasets for general clustering, multiple clustering, and fine-grained clustering. Except Fruit, Card, CUB Birds, and Stanford Cars, other datasets do not provide explicit guideline from their source. We describe all datasets and available guidelines in this section and introduce our prompts to derive guidelines for datasets that do not have ones.

\textit{STL-10}~\cite{coates2011analysis}. STL-10 is an image classification dataset derived from ImageNet, containing 13,000 images across 10 common object categories (airplane, bird, car, cat, deer, dog, horse, monkey, ship, truck). The dataset features 96×96 color images and was designed to evaluate unsupervised feature learning and deep learning methods in scenarios with limited labeled data. The dataset represents general-purpose object categorization with relatively balanced class distributions, making it a standard benchmark for evaluating general clustering methods. Unlike fine-grained datasets, STL-10 focuses on coarse-level semantic distinctions between different object types. Explicit guideline is not available in this dataset.

\textit{CIFAR-10}~\cite{krizhevsky2009learning}. CIFAR-10 is a widely-used image classification dataset consisting of 60,000 32×32 color images across 10 common object categories (airplane, automobile, bird, cat, deer, dog, frog, horse, ship, truck). The dataset is evenly distributed with 6,000 images per class, split into 50,000 training images and 10,000 test images. The low-resolution images and balanced class distribution make CIFAR-10 a standard benchmark for evaluating clustering algorithms on general object categorization tasks. Despite its relatively simple visual characteristics compared to higher-resolution datasets, CIFAR-10 remains challenging for clustering methods due to high intra-class variation and inter-class similarity among certain categories. Explicit guideline is not available in this dataset.

\textit{ImageNet-10}~\cite{deng2009imagenet}. ImageNet-10 is a subset of the large-scale ImageNet dataset, consisting of 13,000 high-resolution images from 10 selected classes. Following existing works in clustering evaluation, we use the specific ImageNet synsets: n02056570 (dog), n02085936 (Maltese dog), n02128757 (lynx), n02690373 (airliner), n02692877 (aircraft carrier), n03095699 (container ship), n04254680 (soccer ball), n04285008 (sports car), n04467665 (trailer truck), and n07747607 (orange). These classes include both general-level categories (e.g., airliner, sports car) and fine-grained distinctions within similar object types (e.g., different dog breeds, various vehicles), creating a challenging clustering scenario that requires semantic understanding at multiple levels of granularity. Explicit guideline is not available in this dataset.

\textit{Fruit}~\cite{hu2017finding}. The Fruit dataset is a multiple clustering dataset. We use all criteria used for multiple clustering (species and color) to construct the guideline.

\textit{Card}~\cite{yao2023augdmc}. The Card dataset is a multiple clustering dataset. We use all criteria used for multiple clustering (number and suits) to construct the guideline.

\textit{CIFAR10-MC}~\cite{yao2024customized}. The CIFAR10-MC dataset is a multiple clustering version of the CIFAR-10 dataset. We use all criteria used for multiple clustering (type and environment) to construct the guideline.

\textit{CUB Birds}~\cite{WahCUB_200_2011}. The Caltech-UCSD Birds-200-2011 (CUB) dataset is a fine-grained visual categorization dataset containing 11,788 images of 200 bird species. The dataset provides rich attribute annotations including visual characteristics such as bill shape, wing color, tail shape, and other morphological features, which we leverage to construct our clustering guideline. Unlike general clustering datasets that focus on coarse-level distinctions between different object types, CUB requires fine-grained discrimination among visually similar bird species based on subtle variations in appearance. This makes CUB particularly challenging for clustering methods, as intra-class variation (e.g., different poses, lighting conditions) can be comparable to inter-class differences between similar species. The availability of detailed attribute annotations makes CUB well-suited for evaluating guideline-driven clustering approaches. 

\textit{Stanford Cars}~\cite{krause20133d}. The Stanford Cars dataset is a fine-grained visual categorization dataset containing 16,185 images of 196 car classes.  Each car class is defined by the combination of Make, Model, and Year (e.g., "2012 Tesla Model S", "2007 BMW M3"), which we use to construct our clustering guideline. The dataset presents unique challenges for fine-grained clustering as cars within the same make and model but different years may have subtle design variations, while cars from different manufacturers can share similar body styles and visual characteristics. Images contain cars from various viewpoints and in different settings, adding to the difficulty of visual discrimination. 

\textit{Stanford Dogs}~\cite{KhoslaYaoJayadevaprakashFeiFei_FGVC2011}. The Stanford Dogs dataset is a fine-grained visual categorization dataset containing 20,580 images across 120 dog breeds from around the world. The dataset encompasses diverse breeds ranging from visually distinct categories (e.g., Chihuahua vs. Saint Bernard) to highly similar breeds that differ primarily in subtle characteristics such as coat texture, ear shape, or body proportions (e.g., different types of terriers or retrievers). Images exhibit significant variation in pose, scale, background clutter, and lighting conditions, making visual feature extraction challenging. The fine-grained nature of dog breed classification requires careful attention to discriminative details that distinguish closely related breeds, making Stanford Dogs particularly demanding for clustering methods. Explicit guideline is not available in this dataset.

\begin{figure}[!t]
    \centering
    \includegraphics[width=1.0\linewidth]{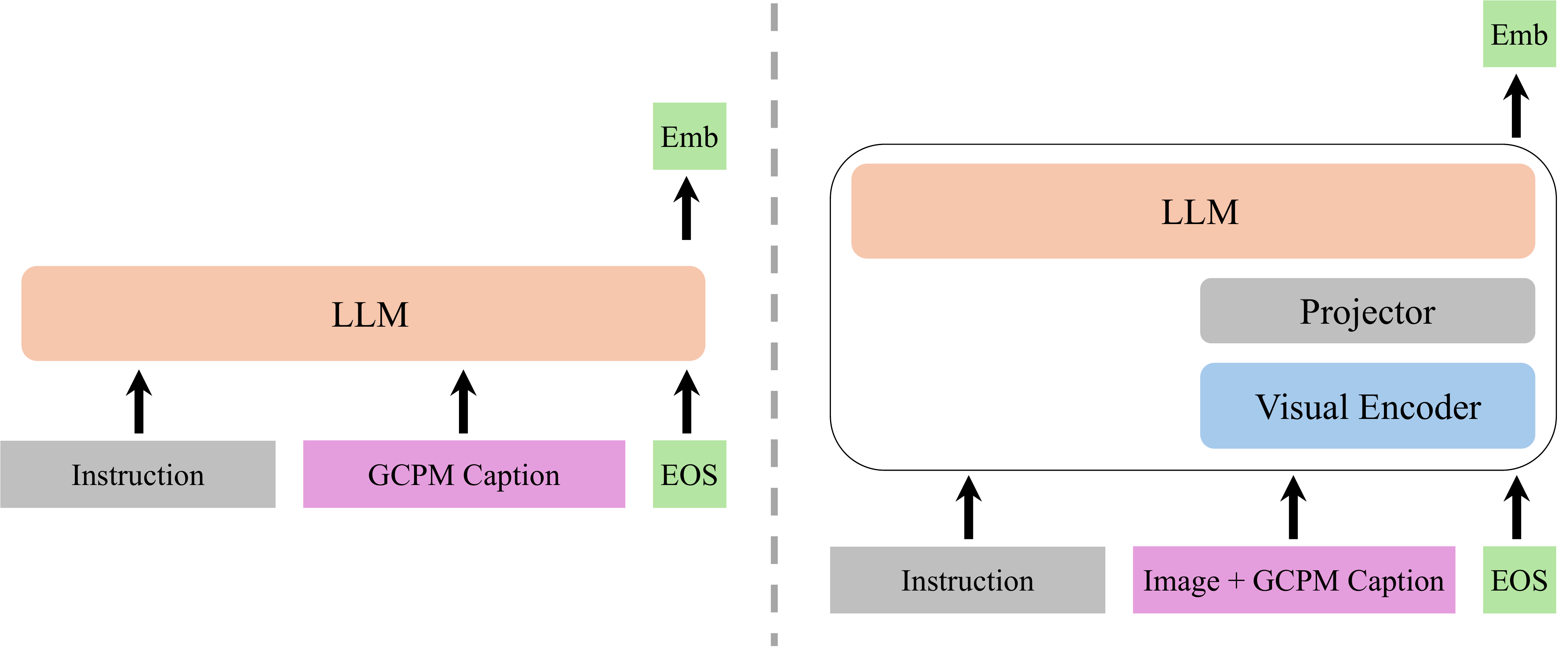}
    \caption{\textbf{Left}: the forwarding process of E5-Mistral. The input contains the instruction, the GCPM caption, and an EOS token, which are concatenated and fed to the LLM. The output EOS token embedding is used for representation. \textbf{Right}: for GME-QWen, the input includes the original image, which are processed to visual tokens by a visual encoder and a projector.}
    \label{fig: forward_process}
\end{figure}
\textit{Oxford Flowers}~\cite{Nilsback08}. The Oxford Flowers dataset is a fine-grained visual categorization dataset containing 8,189 images of 102 flower species commonly found in the United Kingdom. The dataset presents significant challenges for fine-grained clustering due to the high visual similarity among different flower species, where discrimination often relies on subtle differences in petal shape, color patterns, stamen structure, and overall flower morphology. Images exhibit large variations in scale, pose, and illumination, with flowers captured from different angles and in various lighting conditions. Additionally, some species may share similar colors or shapes, requiring careful attention to multiple attributes simultaneously. Explicit guideline is not available in this dataset.

\begin{figure*}
    \centering
    \includegraphics[width=0.85\linewidth]{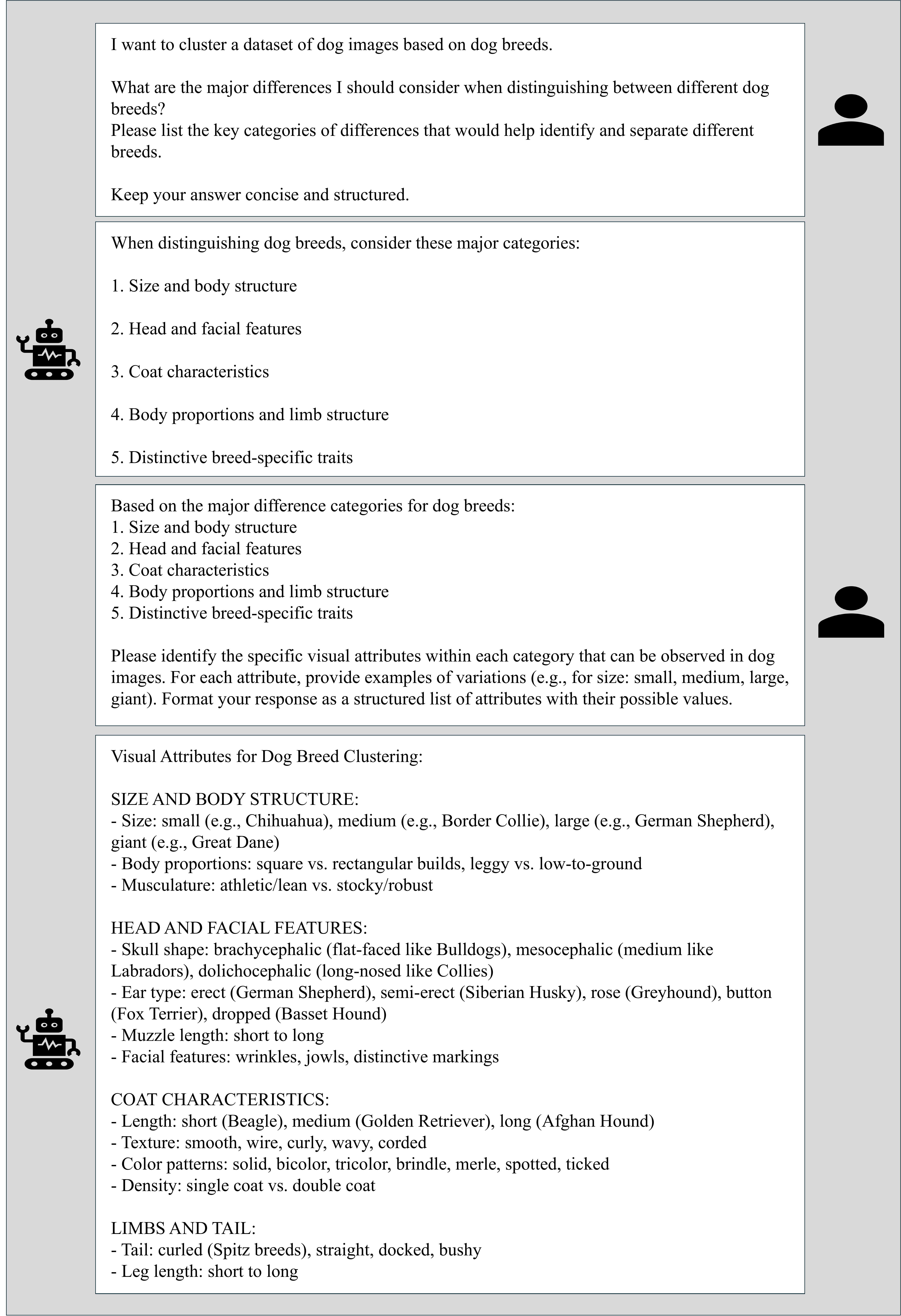}
    \caption{Guideline Generation Process for the Stanford Dogs dataset}
    \label{fig: guideline_gen}
\end{figure*}
\section{Guideline Processing}
\subsection{Guideline Generation}
\label{sec: guideline_gen}
Existing works~\cite{luo2024llm} have explored subpopulation structure discovery within a dataset via LLMs, which can be directly applied to discover key attributes of a dataset. While these methods can be used without any prior knowledge about a dataset, the involvement of LLMs can still be expensive. We consider a more realistic scenario where users typically have a fundamental understanding of a dataset, especially regarding what clustering objective is desired for the dataset. For example, on the Stanford Dogs dataset, users only need to provide the basic clustering objective \textit{dog breeds}. Based on this objective, we introduce a series of heuristic prompts as shown in Figure~\ref{fig: guideline_gen}. The guideline generation process consists of three steps. First, we query the LLM about the major differences when considering possible categories according to the clustering objective. Second, we ask the LLM to derive key visual attributes to consider for clustering. Finally, we  formalize a guideline prompt considering all key visual features as attributes. The three-step generation acts like Chain-of-Thought prompting~\cite{wei2022chain}, leveraging the internal knowledge of LLMs to derive a feasible guideline for clustering.

\subsection{Detailed Guideline}
\label{sec: detailed_guideline}
As mentioned in Section 3.1, we consider a guideline $G$ that contains multiple attributes $A = \{a_1, a_2, \cdots, a_k\} \subseteq G$. Table~\ref{tab: guideline_attributes} shows all attributes considered for each dataset. Based on these attributes, three types of guideline prompts are derived: the GCPM captioning prompts (shown in Tables~\ref{tab: gcpm_gc_prompts}, \ref{tab: gcpm_mc_prompts}, \ref{tab: gcpm_fc_prompts}, \ref{tab: gcpm_lc_prompts}) generate GCPM captions based on all attributes; the instruction-aware embedding prompts (shown in Table~\ref{tab: instruction_aware_prompts}) encode the GCPM captions based on one attribute or a global definition of all attributes; and the MST Traversal prompts (some shown in Table~\ref{tab: merging_prompts} due to space constraints) consider one attribute or a global definition of all attributes for merging a pair of clusters.

\subsection{Guideline and Image Forwarding Process}
\label{sec: forward_process}
We leverage three embedders: Instructor-large~\cite{su2022one}, E5-Mistral~\cite{wang2023improving}, and GME-QWen~\cite{zhang2024gme} for our main experiments. In general, they can be classified into two categories: text-only instruction-aware embedders and multimodal instruction-aware embedders, as shown in Figure~\ref{fig: forward_process}. For Instructor-large and E5-Mistral, the input consists of the GCPM caption and the instruction-aware embedding prompt. For GME-QWen, the input consists of the original image, the GCPM caption, and the instruction-aware embedding prompt.

\subsection{Improving Guideline Accuracy}
\label{sec: improve_guideline}
In this section, we discuss how to improve guideline reliability. From the Bcubed results (Table~\ref{tab: bcubed_results}), we observe that the Bcubed Recall increases while the Precision drops. We argue that this can be attributed to two factors: (1) guidelines are generated in an unsupervised manner without explicit ground truth, potentially introducing ambiguity, and (2) the LLM may occasionally struggle with interpreting subtle guideline differences. Therefore, our improvement goal is to maintain Bcubed Precision while improving Recall. To improve guideline quality, we assume a few-shot supervised setting and use prompt optimization to update the prompt.

\begin{table}[t]
\centering
\caption{Effectiveness of Prompt Optimization using Prompt Agent on the Oxford Flowers dataset.}
\setlength{\tabcolsep}{8pt}
\begin{tabular}{lccc}
\toprule
& NMI & B-Prec. & B-Rec. \\
\midrule
Before Optim. & 88.6 & 90.1 & 67.9 \\
After Optim.  & 90.2 & 95.7 & 65.9 \\
\bottomrule
\end{tabular}
\label{tab: prompt_optim}
\end{table}
We use the Oxford Flowers dataset for experiments. Specifically, we collect 20 false positive pairs, where two clusters that should not be merged are mistakenly merged together. We then use PromptAgent~\cite{wang2023promptagent} with DeepSeek~\cite{liu2024deepseek} as the optimizer to update the prompt. Experimental results are shown in Table~\ref{tab: prompt_optim}. From the experiments, we observe that Bcubed Precision increases together with Recall. This experiment demonstrates that the guideline can be improved when users are not satisfied with its quality, using only very few training samples.

\section{Comparison with Existing Methods}
\label{sec: compare_difference}

\begin{table*}[t]
\centering
\caption{Comparison of key guideline-aware and LLM/VLM-driven clustering methods.}
\setlength{\tabcolsep}{4pt}
\renewcommand{\arraystretch}{1.25}
\scalebox{0.52}{
\begin{tabular}{l|l|l|l|l}
\toprule
\textbf{Paper} & \textbf{Task} & \textbf{Guideline Mechanism} & \textbf{LLM/VLM Role} & \textbf{Limitation/Difference} \\
\midrule

Agentic Graph Traversal (Arxiv 2025) &
Personalized multiple clustering &
User preferences bias graph \& traversal &
MLLM agents expand/merge clusters &
High LLM cost; scaling issues \\

Multi-Sub (NeurIPS 2024) &
Multi-view clustering with user criteria &
GPT-4 proxy words (e.g., \textit{red}, \textit{green}) &
Aligns images to subspaces &
Oversimplifies complex guidelines \\

IC$|$TC (ICLR 2024) &
Image clustering under user text criteria &
User text refined into cluster names &
GPT-4 generates cluster names \& assignments &
Relies on text descriptions; limited fine-grained control \\

LiSA (ACL 2025) &
Topic modeling &
LLM generates candidate topics &
Guides document--topic alignment &
Text-only; not visual \\

ClusterLLM (EMNLP 2023) &
Text clustering &
Triplet/pairwise queries under instruction &
LLM judges similarity \& granularity &
Expensive; not image-focused \\

TAC (ICML 2024) &
Image clustering + external ontology &
WordNet nouns as guidance &
CLIP + noun anchors &
Requires fixed ontology; not user-adaptive \\

TGAICC (Arxiv 2024) &
Multiple valid clusterings &
Different prompts $\rightarrow$ consensus &
LLM generates criteria \& consensus &
Limited efficiency; unstable across prompts \\

ITGC (Arxiv 2025) &
Text-guided image clustering &
Extracts and cluster in concept space &
LLM produces concept-aligned embedding space &
High-quality text needed; weak for fine-grained/hierarchy \\

X-Cluster/TeDeSC (Arxiv 2024) &
Open-ended clustering &
Auto-discovers natural-language criteria &
VLM + LLM propose grouping dimensions &
Lack of control; exploratory, not precise \\

TGIC (EACL 2024) &
Caption-driven clustering &
Captions serve as criteria &
Text model augments embeddings &
Early/simple; weak guideline following \\

CLIP-Cluster (ICCV 2023) &
Attribute-driven face clustering &
Hallucinated CLIP attributes &
CLIP acts as proxy feature extractor &
Not LLM; limited to predefined attributes \\

\midrule
\textbf{Our Clustering Agent} &
General, multi-granularity clustering &
Guideline $\rightarrow$ GCPM $\rightarrow$ MST Traversal &
LLM extracts guideline signals; minimal LLM use &
Training-free, efficient, complex guideline following \\

\bottomrule
\end{tabular}}
\label{tab: comparison_methods}
\end{table*}

\subsection{Guideline-aware and Text-guided Clustering}
\paragraph{Multiple-clustering with user-specified aspects.}
Recent works such as Multi-MaP~\cite{yao2024multi} and Multi-Sub~\cite{yao2024customized} focus on multiple clustering, where a dataset can be partitioned from different perspectives, for example "by color" versus "by species" for fruits. They typically construct proxy spaces or subspaces aligned with user-specified aspects, often using CLIP plus a small number of aspect words as anchors, and then train an end-to-end deep clustering model to produce aspect-specific partitions. While they support user-specified criteria, they (1) require dataset-specific training or fine-tuning, (2) mainly target balanced and relatively small-scale benchmarks such as Fruit, Cards, and CIFAR10-MC, and (3) still assume a small number of discrete aspect choices rather than arbitrary composite guidelines.

\paragraph{Externally guided general image clustering.}
Image Clustering with External Guidance (TAC)~\cite{li2023image} introduces the idea of externally guided clustering by injecting WordNet concepts into the representation space. TAC first builds visual semantic centers, retrieves relevant noun concepts via CLIP, and then jointly distills image and text cluster heads with a mutual-distillation objective. This improves general image clustering without labels, but the "guidance" is limited to a fixed concept vocabulary and cannot express complex multi-attribute business rules or fine-grained operational guidelines. The method is also tied to a specific training pipeline and does not directly extend to multiple clustering, fine-grained clustering, or long-tail settings under a single framework.

\paragraph{Agent-centric personalized multiple clustering.}
Agent-Centric Personalized Multiple Clustering with MLLMs~\cite{chen2025agent} moves closer to agentic guideline following. It uses an MLLM to build an interest-biased similarity graph and then conducts agent-based graph traversal to grow clusters under a specified user interest. The approach enables more explicit user control and uses LLM-based agents to accept or reject membership decisions. However, it still focuses on multiple-clustering style scenarios and relies on training an agent pipeline with GPT-based supervision. Moreover, it assumes a relatively controlled number of clusters and does not target highly long-tailed distributions or automatic cluster discovery in general.

\paragraph{How our method differs.}
Our guideline-driven clustering agent departs from these approaches in several key aspects. First, instead of training a dataset-specific model for each task or aspect, we adopt a \emph{training-free} two-stage hybrid design. Generative Concept Proxy Modeling (GCPM) uses MLLMs plus instruction-aware embedders to produce guideline-aware embeddings from concept proxy captions, without any task-specific fine-tuning. This allows us to reuse the same pipeline across general clustering, multiple clustering, fine-grained clustering, and long-tail e-commerce clustering under a single universal framework.

Second, prior guideline-aware methods generally handle a single criterion per run (for example, "by color" or "by species") or a small enumerated set of views. In contrast, our formulation treats a guideline as a composition of attributes and explicitly models composite criteria such as "group by brand and intended activity" or domain-specific SOP-style rules. The concept proxy step disentangles these attributes, and the subsequent instruction-aware encoding enables flexible recombination of attributes within the embedding space.

Third, while methods like TAC and multiple-clustering approaches assume either known cluster numbers or balanced distributions, we explicitly tackle scenarios where the number of clusters is unknown and the distribution is highly long-tailed. We use HDBSCAN to obtain conservative small clusters and then apply MST-based LLM Traversal to semantically merge clusters according to the guideline. This hybrid strategy allows us to handle automatic cluster discovery and large numbers of tiny clusters that conventional multiple-clustering methods and externally guided approaches do not address.
Overall, existing guideline-aware and multiple-clustering methods either require substantial training, support only simple criteria, or target a narrow set of scenarios, whereas our universal guideline-driven agent is training-free, supports composite guidelines, and covers GC, MC, FC, and long-tail clustering within one coherent framework.

\subsection{LLM-based Clustering}
A second line of related work uses large language models more directly in the clustering process, either as supervisors, similarity oracles, or agentic controllers. These methods demonstrate the semantic power of LLMs but typically do not address complex guideline-driven image clustering in a universal way.

\paragraph{LLMs as clustering supervisors or guides.}
ClusterLLM~\cite{zhang2023clusterllmlargelanguagemodels} treats an API-only LLM as a teacher that guides unsupervised text clustering. It uses triplet queries to refine a small local embedder to better align with an instruction such as "cluster by topic" or "cluster by sentiment", and then queries the LLM again on pairwise decisions to select an appropriate level of a hierarchical clustering tree. Large Language Models Enable Few-Shot Clustering~\cite{viswanathan2023largelanguagemodelsenable} similarly leverages LLMs to produce cluster descriptions and document assignments from few labeled examples, largely in text domains. Both methods showcase how LLMs can steer clustering criteria and granularity, but they operate mainly on textual data, rely on repeated LLM interactions over a large set of examples, and do not handle rich visual guidelines or the diverse image clustering regimes we consider.
LLM-Guided Semantic-Aware Clustering for Topic Modeling (LiSA)~\cite{liu2025llm} further integrates LLMs into topic modeling by generating topic words and descriptions per document, clustering both documents and topics, and then using an LLM in a conflict resolution module. Its focus is on aligning topic and document for text corpora, not on supporting user-specified guidelines or visual semantics.

\paragraph{LLM-based clustering in graphs and intents.}
Recent works such as MARK~\cite{fu2025mark}, SPILL~\cite{lin2025spill}, and ZeroDL~\cite{jo2024zerodl} extend LLM-based clustering to text-attributed graphs, intent discovery, and zero-shot distribution learning. MARK uses multi-agent collaboration with ranking guidance on text-attributed graphs, SPILL focuses on domain-adaptive intent clustering with LLM-based selection and pooling, and ZeroDL formulates zero-shot distribution learning for text clustering via LLMs. These methods again show that LLMs can provide powerful semantic guidance but are designed for textual or graph data, and they typically assume relatively well-behaved cluster structures rather than severe long-tail distributions.
Information-Theoretic Generative Clustering of Documents (ITGC)~\cite{du2024informationtheoreticgenerativeclusteringdocuments} represents another direction that combines information-theoretic objectives with LLM components for document clustering. ITGC improves text clustering quality by modeling generative relationships and mutual information, but it remains confined to document data and does not address image semantics, guideline composition, or agentic control across multiple clustering tasks.

\paragraph{LLM-based image clustering.}
On the vision side, Image Clustering Conditioned on Text Criteria (IC$|$TC)~\cite{kwon2023image} is the most relevant LLM-based image clustering work. It uses a VLM to produce criterion-specific image descriptions, then asks an LLM to propose cluster names and assign each description to one of K clusters given a user-provided text criterion. This enables training-free text-guided clustering on images but has several important limitations. First, it still assumes a single concrete criterion per run, such as "cluster by action" or "cluster by location", and does not support multi-attribute guidelines or complex constraints. Second, it assumes the number of clusters K is known or searched over in a simple fashion, and it is not tailored to long-tail settings with many tiny clusters. Third, it performs LLM-based assignment over all samples, which leads to high LLM cost when scaling to large datasets or multiple successive clustering tasks.

\paragraph{How our method differs.}
Our approach can be seen as a hybrid between LLM-guided clustering and deep clustering, but with a focus on \emph{universal guideline-driven image clustering}. Compared to ClusterLLM, LiSA, and other text-oriented LLM clustering methods, we operate directly on vision tasks (GC, MC, FC, and LC) and treat guidelines as composable multi-attribute descriptions rather than single textual perspectives. We also avoid repeated fine-tuning of embedders or training of new prediction networks, remaining fully training-free.

Relative to IC$|$TC and other LLM-based image clustering approaches, our framework introduces two key advances. First, GCPM disentangles guideline attributes through concept proxy captions and instruction-aware embeddings, which allows us to reuse the same embeddings across different criteria and composite guidelines, and to plug them into both standard clustering algorithms and the MST-based traversal stage. Second, MST-based LLM Traversal uses HDBSCAN to generate conservative small clusters and then performs LLM queries only on a carefully selected subset of cluster pairs along a minimum spanning tree. This design reduces the number of LLM calls from quadratic in the number of clusters to approximately $O(M \log M)$ in expectation, while still enabling guideline-aware semantic merging. In contrast, IC$|$TC and many LLM-based clustering methods invoke the LLM on a large fraction of examples or cluster pairs, leading to higher computational cost.

Finally, most existing LLM-based clustering works assume either known cluster numbers or relatively balanced clusters. By combining guideline-aware embeddings with HDBSCAN and LLM-based merging, our method explicitly targets settings where the cluster number is unknown and clusters are extremely imbalanced, such as the long-tail ABO e-commerce dataset with thousands of tiny clusters. This capability is not addressed by prior LLM-based clustering or agentic frameworks and is crucial for deploying guideline-driven image clustering in realistic scenarios.

\subsection{General Clustering Baselines}
We compare our guideline-driven agent with a diverse set of state-of-the-art general deep clustering methods on CIFAR-10, STL-10, and ImageNet-10, as summarized in Table~2.\footnote{See the main text for dataset definitions.} All baselines assume that the ground-truth number of clusters is known and train task-specific models on each dataset, while our method remains training-free and can operate both with and without this prior.
\paragraph{Classical and constrained deep clustering.}
Cop-KMeans (Cop-KMS.)~\cite{wagstaff2001constrained} extends K-Means with pairwise constraints. IIC~\cite{ji2019invariant} and DCCM~\cite{wu2019deep} are early deep clustering methods that optimize instance consistency or mutual information between augmented views. These methods learn dataset-specific encoders and clustering heads, usually optimized per dataset and metric.
\paragraph{Self-supervised representation learning with clustering.}
A large group of baselines follows the recent paradigm of first learning general-purpose representations via self-supervised objectives, then performing K-Means in the embedding space. BYOL~\cite{grill2020bootstrap}, MiCE~\cite{tsai2020mice}, PCL\cite{li2020prototypical}, PICA\cite{huang2020deep}, SCAN~\cite{van2020scan}, FixMatch~\cite{sohn2020fixmatch}, CC~\cite{li2021contrastive}, GCC~\cite{zhong2021graph}, IDFD~\cite{tao2021clustering}, NMM~\cite{dang2021nearest}, TCC~\cite{shen2021you}, ProPos~\cite{huang2022learning}, SPICE~\cite{niu2022spice}, TCL~\cite{li2022twin}, CoNR~\cite{yu2023contextually}, DMICC~\cite{li2023dual}, SeCu~\cite{qian2023stable}, and DivClust~\cite{metaxas2023divclust} all fall into this category. They typically rely on heavy pretraining and then optimize clustering quality through contrastive, mutual-information, or distribution-regularization losses. Many of them further introduce specialized training objectives for cluster separation or diversity, which improves performance but ties the method to the particular dataset and task.

For all methods above, we either adopt reported numbers from their original papers or follow their publicly released implementations and recommended hyperparameters under the standard GC evaluation protocol (ACC, NMI, and ARI) on CIFAR-10, STL-10, and ImageNet-10. Our method shares the same evaluation protocol but does not perform any fine-tuning on the target datasets.

\subsection{Multiple Clustering Baselines}
Multiple clustering (MC) benchmarks require algorithms to discover several valid partitions of the same dataset according to different criteria (e.g., Fruit by color or species, Cards by number or suit, CIFAR10-MC by object type or environment). Following prior work, we evaluate on Fruit, Cards, and CIFAR10-MC using NMI and RI for each criterion and also report the averaged score per dataset in Table~\ref{tab: mc_results}. All baselines assume the true number of clusters is known and train models specifically tailored to these MC scenarios, whereas our method uses a guideline-driven agent that can be instantiated with any textual clustering specification.

\paragraph{Early multiple-clustering methods.}
MSC~\cite{hu2017finding} and MCV~\cite{guerin2018improving} are early methods for discovering alternative partitions through feature reweighting or multi-view embeddings. ENRC~\cite{miklautz2020deep} and iMClusts~\cite{ren2022diversified} refine this line by learning disentangled embeddings that encourage different clustering heads to focus on complementary aspects of the data. These methods are effective on small-scale MC datasets but still rely on explicit training and a fixed number of predefined clusters.

\paragraph{Deep multiple clustering with augmented or diffusion-based representations.}
AugDMC~\cite{yao2023augdmc} and DDMC~\cite{yao2024dual} introduce stronger deep backbones and augmentation- or diffusion-based regularization to improve the robustness of multiple clusterings. They optimize several heads jointly to encourage diversity between perspectives while maintaining high consistency with each ground-truth partition. The models are trained per dataset and per set of criteria and thus cannot be easily reused for new guidelines without retraining.

\paragraph{State-of-the-art proxy-based multiple clustering.}
Multi-MaP~\cite{yao2024multi} and Multi-Sub~\cite{yao2024customized} represent the current state of the art for MC. They introduce proxy-based modules that map images to concept anchors or subspaces aligned with user-specified criteria, then perform clustering in each learned representation. While these methods can flexibly switch between a small set of predefined clustering dimensions, they still require extensive training on labeled or pseudo-labeled data for each dataset and assume that all criteria are known ahead of time.
In contrast, our guideline-driven agent directly takes natural language descriptions of the desired criterion (for example, "group by card number" or "group by card suit") and uses the same GCPM embeddings and MST-based LLM Traversal across all MC tasks without retraining. This enables us to support new or composite guidelines beyond the few perspectives that existing MC baselines are designed for, while still achieving competitive or superior performance on the standard Fruit, Cards, and CIFAR10-MC benchmarks.

\subsection{Fine-grained Clustering Baselines}
Fine-grained clustering (FC) benchmarks focus on distinguishing visually similar subcategories, such as bird species, dog breeds, car models, or flower species. We evaluate on CUB Birds, Stanford Dogs, Stanford Cars, and Oxford Flowers using ACC and NMI, following prior work. The compared baselines in Table~4 cover both representation-learning approaches and generative clustering methods. Similar to GC and MC, all FC baselines rely on dataset-specific training and, except for UFCL, assume the number of clusters is known, while our method remains training-free and can operate without that prior.
\paragraph{Representation-learning baselines.}
IIC~\cite{ji2019invariant} extends invariant information maximization to FC settings; SimCLR~\cite{chen2020simple} and MoCo~\cite{he2020momentum} are contrastive self-supervised representation learners whose embeddings are clustered to form categories. SCAN~\cite{van2020scan} and SeCu~\cite{qian2023stable} further adapt self-supervised models to clustering by adding clustering-specific heads and objectives. These approaches show that strong representations help FC, but they still treat clustering as a purely geometric problem in the embedding space and cannot incorporate textual guidelines.
\paragraph{Generative and GAN-based fine-grained clustering.}
A second group of baselines focuses on generative modeling for fine-grained discovery. InfoGAN~\cite{chen2016infogan}, FineGAN~\cite{singh2019finegan}, MixNMatch~\cite{li2020mixnmatch}, OneGAN~\cite{benny2020onegan}, SD~\cite{rombach2022high}, and C3-GAN~\cite{kim2021contrastive} learn generative models whose latent codes are encouraged to align with semantic factors, then interpret these codes as cluster assignments. DiFiC~\cite{yang2024dific} leverages diffusion models for fine-grained semantics. These methods achieve strong performance when trained on each dataset but require substantial computation and bespoke architectures, and they do not provide a direct mechanism to encode user-defined guidelines into the clustering process.
\paragraph{Unified fine-grained clustering.}
UFCL~\cite{wang2023bridge} is a recent unified fine-grained clustering framework that aims to handle several datasets within a single model. It still requires training on fine-grained data and assumes cluster numbers are known in our evaluation. Our method differs in two important aspects: (i) we incorporate textual guidelines to define the target granularity or attributes of interest, and (ii) we remain entirely training-free, combining guideline-aware GCPM embeddings with MST-based LLM Traversal to refine clusters. Despite using conservative prompts for FC to avoid over-merging visually similar categories, our model still achieves state-of-the-art or highly competitive results on CUB Birds, Stanford Dogs, Stanford Cars, and Oxford Flowers, demonstrating that guideline-driven clustering scales even to demanding fine-grained scenarios.

\section{More Ablations}
\label{sec: more_ablations}
\subsection{Hyperparameter Analysis}
In main experiments, we use min\_cluster\_size=2 for HDBSCAN and K=5 for Top-K samples representing a cluster in the MST Traversal. We conducted hyperparameter analysis on Card-Number for these parameters as shown in Table~\ref{tab: hyper_param}.
\begin{table}[t]
\centering
\caption{Ablation on hyperparameter settings. We report NMI scores for GME-QWen on Card-Number.}
\setlength{\tabcolsep}{6pt}
\begin{tabular}{lc}
\toprule
Hyperparameter Settings & Results \\
\midrule
Top-K (K=3) & 70.7 \\
Top-K (K=5) & 72.1 \\
Top-K (K=7) & 72.4 \\
\midrule
HDBS. min\_cluster\_size=2 & 50.3 \\
HDBS. min\_cluster\_size=3 & 55.4 \\
HDBS. min\_cluster\_size=4 & 56.7 \\
\bottomrule
\end{tabular}
\label{tab: hyper_param}
\end{table}
In experiments, we observe that the K=5 achieves a relative good balance for performance and efficiency. We found that the performance varies for changes in min\_cluster\_size. Though large values seem to achieve better performance, we maintain min\_cluster\_size=2 because of the philosophy that at least two clusters are enough to form a cluster other than singletons. It can be applied to all dataset we tested. 

\subsection{GCPM with Other Clustering Algorithms}
We tested GCPM embeddings using GME-QWen with other clustering algorithms, including Agglomerative Clustering (AC), Spectral Clustering (SC), and DBSCAN, as shown in Table~\ref{tab: diff_clus_algo}.
\begin{table}[t]
\centering
\caption{Clustering NMI results for GCPM embeddings using GME-QWen on Card-Number. SC: Spectral Clustering; AC: Agglomerative Clustering.}
\setlength{\tabcolsep}{8pt}
\begin{tabular}{lc}
\toprule
& Card-Number \\
\midrule
AC      & 84.7 \\
SC      & 81.1 \\
DBSCAN  & 44.4 \\
\bottomrule
\end{tabular}
\label{tab: diff_clus_algo}
\end{table}
Note that we provide the number of clusters for both AC and SC. We observe that their performance is comparable with K-Means.

\subsection{MST Traversal with Proprietary LLMs}
We tested MST Traversal using a proprietary and more advanced LLM, Claude-3.5-Sonnet. Experimental results are shown in Table~\ref{tab: claude_results}.
\begin{table}[t]
\centering
\caption{Evaluation of Proprietary LLMs on Card-Number and Stanford Cars.}
\setlength{\tabcolsep}{8pt}
\begin{tabular}{l|cc|cc}
\toprule
& \multicolumn{2}{c|}{Card-Number} & \multicolumn{2}{c}{Stanford Cars} \\
\cmidrule(lr){2-3} \cmidrule(lr){4-5}
& NMI & RI & NMI & ARI \\
\midrule
QWen-VL & 72.1 & 95.1 & 80.9 & 33.2 \\
Claude  & 79.4 & 96.3 & 82.2 & 41.0 \\
\bottomrule
\end{tabular}
\label{tab: claude_results}
\end{table}
In general, we observe that Claude achieves better performance than QWen-VL. We argue the performance improvement is because of the stronger reasoning capability of Claude.

\subsection{Using MST Traversal for Existing Baselines}
Most existing baselines directly use K-Means for clustering. We demonstrate that MST Traversal can improve their performance when applied to HDBSCAN, as shown in Table~\ref{tab: mst_on_multi_sub}.
\begin{table}[t]
\centering
\caption{Effect of MST Traversal on Multi-Sub.}
\setlength{\tabcolsep}{8pt}
\begin{tabular}{lcc}
\toprule
& NMI & RI \\
\midrule
Multi-Sub         & 34.2 & 80.7 \\
+MST Traversal    & 37.4 & 85.9 \\
\bottomrule
\end{tabular}
\label{tab: mst_on_multi_sub}
\end{table}
We reimplement Multi-Sub and test it using HDBSCAN. Upon clustering by HDBSCAN, we apply our MST Traversal to improve the clustering result. Experiments again validate the effectiveness of the MST Traversal.

\section{Reproduction and Limitations}
\paragraph{Reproduction} Complete codes and the processed ABO-LC dataset will be released upon acceptance.

\paragraph{Limitations and Future Directions}
The effectiveness of our approach depends on the quality and specificity of input guidelines. When guidelines are ambiguous or incomplete, the clustering results may not fully align with user intentions. To address this, we have explored two complementary strategies: (1) heuristic prompting techniques that help users articulate their clustering objectives more clearly, and (2) prompt optimization methods (Section~\ref{sec: improve_guideline}) that refine guidelines using minimal feedback. Our experiments (Table~\ref{tab: prompt_optim}) demonstrate that even with limited supervision (20 samples), prompt optimization can significantly improve both precision and recall, suggesting a practical path for iterative refinement in real-world deployments.

Although our MST-based traversal algorithm substantially reduces LLM invocations compared to naive approaches (Section~\ref{sec: runtime_complexity}, Figure~\ref{fig: merge_rate}), the method still requires multiple LLM calls for complex datasets with thousands of initial clusters. For extremely large-scale applications, exploring more efficient semantic reasoning mechanisms (such as caching strategies, hierarchical merging, or hybrid similarity metrics) represents a promising direction. Nevertheless, our current design achieves a favorable balance between semantic accuracy and computational efficiency, as evidenced by our $O(N \log N)$ complexity analysis.

\begin{table*}[t]
\centering
\caption{Guideline attributes used for each dataset.}
\setlength{\tabcolsep}{4pt}
\renewcommand{\arraystretch}{1.15}
\begin{tabular}{l|p{0.8\textwidth}}
\toprule
Dataset & Guideline Attributes \\
\midrule
CIFAR-10 & Object category; Living status; Habitat/Environment; Locomotion; Body type; Size range; Surface covering; Primary color; Key distinguishing features \\
\midrule
STL-10 & Object type; Main shape; Primary environment; Movement type; Size category; Surface texture; Main color; Key features \\
\midrule
ImageNet-10 & Object type; Living status; Natural habitat; Primary function; Physical form; Size category; Surface texture; Primary color; Key features \\
\midrule
Fruit & Species; Color \\
\midrule
Cards & Number; Suits \\
\midrule
CIFAR10-MC & Object category; Living status; Habitat/Environment; Locomotion; Body type; Size range; Surface covering; Primary color; Key distinguishing features \\
\midrule
CUB Birds & Bill shape; Wing color; Upperparts color; Underparts color; Breast pattern; Back color; Tail shape; Upper tail color; Head pattern; Breast color; Throat color; Eye color; Bill length; Forehead color; Under tail color; Nape color; Belly color; Wing shape; Size; Shape; Back pattern; Tail pattern; Belly pattern; Primary color; Leg color; Bill color; Crown color; Wing pattern \\
\midrule
Stanford Cars & Make; Model; Year \\
\midrule
Stanford Dogs & Size; Body proportions; Musculature; Skull shape; Ear type; Muzzle length; Facial features; Length; Texture; Color patterns; Density; Tail; Leg length \\
\midrule
Oxford Flowers & Petal colors; Color distribution; Center/stamen color contrasts with petals; Petal shape; Number of petals; Arrangement of petals; Flower head structure; Petal texture; Surface patterns; Size relationships between flower parts; Density; Complexity; Growth pattern; Visible foliage characteristics \\
\midrule
ABO-LC & Brand; Color; Item id; Model name; Product type; Style\\
\bottomrule
\end{tabular}
\label{tab: guideline_attributes}
\end{table*}
\begin{table*}[t]
\centering
\caption{GCPM Caption Prompts for General Clustering Datasets.}
\setlength{\tabcolsep}{4pt}
\renewcommand{\arraystretch}{1.12}
\begin{tabular}{l|p{0.8\textwidth}}
\toprule
Dataset & GCPM Caption Prompt \\
\midrule

CIFAR-10 &
``Describe the object in the image with a specific focus on required criteria for general object recognition.

You should strictly follow the output format below:

OUTPUT FORMAT:
- Description: Provide a general description of the main object in the image here.
- Object category: Provide the broad category here.
- Living status: Provide the living status here.
- Habitat/Environment: Provide the typical habitat or environment here.
- Locomotion: Provide how it moves here.
- Body type: Provide the body structure here.
- Size range: Provide the typical size here.
- Surface covering: Provide the main surface here.
- Primary color: Provide the dominant color here.
- Key distinguishing features: Describe 2-3 most distinctive visual features that help identify this specific object type.'' \\
\midrule
STL-10 &
``Describe the object in the image with a specific focus on required criteria for general object recognition.

You should strictly follow the output format below:

OUTPUT FORMAT:
- Description: Provide a general description of the main object in the image here.
- Object type: Provide the type of object here.
- Main shape: Provide the overall shape here.
- Primary environment: Provide the typical environment here.
- Movement type: Provide how it typically moves here.
- Size category: Provide the relative size here.
- Surface texture: Provide the main surface texture here.
- Main color: Provide the dominant color here.
- Key features: Describe 2-3 most distinctive features that help identify this object.'' \\
\midrule
ImageNet-10 &
``Describe the object in the image with a specific focus on required criteria for general object recognition.

You should strictly follow the output format below:

OUTPUT FORMAT:
- Description: Provide a general description of the main object in the image here.
- Object type: Provide the broad type here.
- Living status: Provide the living status here.
- Natural habitat: Provide the natural environment here.
- Primary function: Provide the main function here.
- Physical form: Provide the physical structure here.
- Size category: Provide the typical size here.
- Surface texture: Provide the main surface here.
- Primary color: Provide the dominant color here.
- Key features: Describe 2-3 most distinctive features that help identify this specific object.'' \\
\bottomrule
\end{tabular}
\label{tab: gcpm_gc_prompts}
\end{table*}
\begin{table*}[t]
\centering
\caption{GCPM Caption Prompts for Multiple Clustering Datasets.}
\setlength{\tabcolsep}{4pt}
\renewcommand{\arraystretch}{1.12}
\begin{tabular}{l|p{0.85\textwidth}}
\toprule
Dataset & GCPM Caption Prompt \\
\midrule
Fruit &
``Describe the fruit in the image with a specific focus on required criteria.

You should strictly follow the output format below:

OUTPUT FORMAT:
- Description: Provide a general description of the fruit here.
- Color: Provide the color of the fruit via its hex color code here. Please do not include the exact color because it may be misleading.
- Species: Provide the general species of the fruit here. Do not output Undefined or Unclear. If you cannot decide, try your best to guess one.'' \\
\midrule
Cards &
``Describe the poker card in the image with a specific focus on required criteria.

You should strictly follow the output format below:

OUTPUT FORMAT:
- Description: Provide a general description of the card here.
- Suit: Provide the suit of the card here.
- Number: Provide the number of the card here.'' \\
\midrule
CIFAR10-MC &
``Describe the object in the image with a specific focus on required criteria for general object recognition.

You should strictly follow the output format below:

OUTPUT FORMAT:
- Description: Provide a general description of the main object in the image here.
- Object category: Provide the broad category here.
- Living status: Provide the living status here.
- Habitat/Environment: Provide the typical habitat or environment here.
- Locomotion: Provide how it moves here.
- Body type: Provide the body structure here.
- Size range: Provide the typical size here.
- Surface covering: Provide the main surface here.
- Primary color: Provide the dominant color here.
- Key distinguishing features: Describe 2-3 most distinctive visual features that help identify this specific object type.'' \\
\bottomrule
\end{tabular}
\label{tab: gcpm_mc_prompts}
\end{table*}
\begin{table*}[t]
\centering
\caption{GCPM Caption Prompts for Fine-grained Clustering Datasets.}
\setlength{\tabcolsep}{4pt}
\renewcommand{\arraystretch}{1.12}
\begin{tabular}{l|p{0.82\textwidth}}
\toprule
Dataset & GCPM Caption Prompt \\
\midrule
CUB Birds &
``Describe the bird in the image with a specific focus on required criteria.

You should strictly follow the output format below:

OUTPUT FORMAT:
- Description: Provide a general description of the bird here.
- Has bill shape: Provide the bill shape of the bird here.
- Has wing color: Provide the wing color of the bird here.
- Has upperparts color: Provide the upperparts color of the bird here.
- Has underparts color: Provide the underparts color of the bird here.
- Has breast pattern: Provide the breast pattern of the bird here.
- Has back color: Provide the back color of the bird here.
- Has tail shape: Provide the tail shape of the bird here.
- Has upper tail color: Provide the upper tail color of the bird here.
- Has head pattern: Provide the head pattern of the bird here.
- Has breast color: Provide the breast color of the bird here.
- Has throat color: Provide the throat color of the bird here.
- Has eye color: Provide the eye color of the bird here.
- Has bill length: Provide the bill length of the bird here.
- Has forehead color: Provide the forehead color of the bird here.
- Has under tail color: Provide the under tail color of the bird here.
- Has nape color: Provide the nape color of the bird here.
- Has belly color: Provide the belly color of the bird here.
- Has wing shape: Provide the wing shape of the bird here.
- Has size: Provide the size of the bird here.
- Has shape: Provide the shape of the bird here.
- Has back pattern: Provide the back pattern of the bird here.
- Has tail pattern: Provide the tail pattern of the bird here.
- Has belly pattern: Provide the belly pattern of the bird here.
- Has primary color: Provide the primary color of the bird here.
- Has leg color: Provide the leg color of the bird here.
- Has bill color: Provide the bill color of the bird here.
- Has crown color: Provide the crown color of the bird here.
- Has wing pattern: Provide the wing pattern of the bird here.'' \\
\midrule
Stanford Cars &
``Describe the car in the image with a specific focus on required criteria.

You should strictly follow the output format below:

OUTPUT FORMAT:
- Description: Provide a general description of the car here.
- Make: Provide the make of the car here.
- Model: Provide the model of the car here.
- Year: Provide the year of the car here.'' \\
\midrule
Stanford Dogs &
``Describe the dog in the image with a specific focus on required criteria.

You should strictly follow the output format below:

OUTPUT FORMAT:
- Description: Provide a general description of the dog here.
- Size: Provide the size of the dog here.
- Body proportions: Provide the body proportions of the dog here.
- Musculature: Provide the musculature of the dog here.
- Skull shape: Provide the skull shape of the dog here.
- Ear type: Provide the ear type of the dog here.
- Muzzle length: Provide the muzzle length of the dog here.
- Facial features: Provide the facial features of the dog here.
- Coat length: Provide the coat length of the dog here.
- Coat texture: Provide the coat texture of the dog here.
- Color patterns: Provide the color patterns of the dog here.
- Coat density: Provide the coat density of the dog here.
- Tail: Provide the tail characteristics of the dog here.
- Leg length: Provide the leg length of the dog here.'' \\
\midrule
Oxford Flowers &
``Describe the flower in the image with a specific focus on required criteria.

You should strictly follow the output format below:

OUTPUT FORMAT:
- Description: Provide a general description of the flower here.
- Petal colors: Provide the petal colors of the flower here.
- Color distribution: Provide the color distribution of the flower here.
- Center/stamen color: Provide the center/stamen color of the flower here.
- Petal shape: Provide the petal shape of the flower here.
- Number of petals: Provide the number of petals of the flower here.
- Arrangement of petals: Provide the arrangement of petals of the flower here.
- Flower head structure: Provide the flower head structure of the flower here.
- Petal texture: Provide the petal texture of the flower here.
- Surface patterns: Provide the surface patterns of the flower here.
- Size relationships between flower parts: Provide the size relationships between flower parts of the flower here.
- Petal density: Provide the petal density of the flower here.
- Bloom complexity: Provide the bloom complexity of the flower here.
- Growth pattern: Provide the growth pattern of the flower here.
- Visible foliage characteristics: Provide the visible foliage characteristics of the flower here.'' \\
\bottomrule
\end{tabular}
\label{tab: gcpm_fc_prompts}
\end{table*}
\begin{table*}[t]
\centering
\caption{GCPM Caption Prompts for Long-tailed Clustering Dataset.}
\setlength{\tabcolsep}{4pt}
\renewcommand{\arraystretch}{1.12}
\begin{tabular}{l|p{0.82\textwidth}}
\toprule
Dataset & GCPM Caption Prompt \\
\midrule
ABO-LC &
``Item Name:
{item\_name}

Describe the e-commerce product based on its image and item name with a specific focus on required criteria.

You should strictly follow the output format below:

OUTPUT FORMAT:
- Description: Provide a general description of the product here.
- Brand: Provide the brand of the e-commerce product here, which might be inferred from the item name.
- Model Name: Provide the model name of the e-commerce product here, which might be inferred from the item name.
- Product Type: Provide the product type of the e-commerce product here, which might be inferred from the item name.
- Style: Provide the style of the e-commerce product here, which might be inferred from the item name.'' \\
\bottomrule
\end{tabular}
\label{tab: gcpm_lc_prompts}
\end{table*}
\begin{table*}[t]
\centering
\caption{Instruction-Aware Embedding Prompts for Each Dataset.}
\setlength{\tabcolsep}{4pt}
\renewcommand{\arraystretch}{1.12}
\begin{tabular}{l|p{0.78\textwidth}}
\toprule
Dataset & Instruction-Aware Embedding Prompt \\
\midrule

CIFAR-10 &
Identify the object type based on the image description considering its physical characteristics, habitat, and functional properties. \\
\midrule
STL-10 &
Identify the category of the object based on the image description focusing on distinctive visual and functional characteristics. \\
\midrule
ImageNet-10 &
Identify the specific object category based on the image description considering its physical characteristics, function, and distinctive features. \\
\midrule
Fruit &
Identify the species of the fruit based on the image description. \\
& Identify the color of the fruit based on the image description. \\
\midrule
Cards &
Identify the number of the card based on the image description. \\
& Identify the suit of the card based on the image description. \\
\midrule
CIFAR10-MC &
Identify the type of the object based on the image description. \\
& Identify the environment where the object is based on the image description. \\
\midrule
CUB Birds &
Identify the species of the bird based on the image description. \\
\midrule
Stanford Cars &
Identify the type of the car based on the image description. \\
\midrule
Stanford Dogs &
Identify the breed of the dog based on the image description. \\
\midrule
Oxford Flowers &
Identify the category of the flower based on the image description. \\
\midrule
ABO-LC & Identify the variation of the product based on the description focusing on its brand, model name, product type, and style \\

\bottomrule
\end{tabular}
\label{tab: instruction_aware_prompts}
\end{table*}
\begin{table*}[t]
\centering
\caption{LLM Merging Prompts Used in MST Traversal for Cards conditioned on Number and Stanford Cars.}
\setlength{\tabcolsep}{4pt}
\renewcommand{\arraystretch}{1.12}
\begin{tabular}{l|p{0.78\textwidth}}
\toprule
Dataset & LLM Merging Prompt in MST \\
\midrule

Cards-Number &
``You are an agent for poker card image clustering. Your task is to determine whether two clusters of poker card images should be merged based on their representative images or image descriptions.

(merge\_rules)
    1. Please observe the main value of the cards for each cluster. If two clusters have the same value of cards, they should be merged. Otherwise, they should not be merged.
(/merge\_rules)'' \\
\\[-2mm]
\midrule
Stanford Cars &
``You are an agent for car image clustering. Your task is to determine whether two clusters of car images should be merged based on their representative images or image descriptions.

(merge\_rules)
    1. Two clusters can be merged only if their images show cars with the same make, model, and year. Otherwise, they should not be merged.
    2. ALL of the following features MUST match EXACTLY between clusters:
        - Make (manufacturer brand)
        - Model (specific car model name)
        - Year (production year)
    3. ZERO TOLERANCE POLICY:
        - If ANY of these three features differ, DO NOT MERGE
        - Different years of the same make and model should NOT be merged
        - Different models from the same make should NOT be merged
        - Different trims or variants should be evaluated based on whether they are considered the same model
    4. General appearance similarities are NOT sufficient for merging – the make, model, and year must match exactly.
(/merge\_rules)'' \\
\bottomrule
\end{tabular}
\label{tab: merging_prompts}
\end{table*}